\documentclass[10pt, journal]{IEEEtran}
\usepackage[utf8]{inputenc} 
\DeclareUnicodeCharacter{202F}{ } 

\IEEEoverridecommandlockouts
\usepackage{graphicx}
\usepackage{subcaption}
\usepackage{amsmath,amsthm,amsfonts,amssymb}
\usepackage{cite}
\usepackage{bm}
\usepackage{bbm}
\usepackage{url}
\usepackage{array}
\usepackage{color}
\usepackage{multirow}
\usepackage{booktabs}
\usepackage[table,xcdraw]{xcolor}
\usepackage{enumerate}
\usepackage{epstopdf}
\usepackage{tabularx}
\usepackage{algorithm}
\usepackage[english]{babel}
\usepackage{algpseudocode}
\usepackage{makecell}
\usepackage{tablefootnote} 
\usepackage{threeparttable}

\algrenewcommand{\algorithmiccomment}[1]{\hfill// #1}

\allowdisplaybreaks[4]

\begin{document}
\title{\makebox[\linewidth]{\parbox{\dimexpr\textwidth+0cm\relax}{\centering A Hybrid TDMA/CSMA Protocol for Time-Sensitive Traffic in Robot Applications}}}

\author{
Shiqi~Xu$^*$, 
Lihao~Zhang$^*$, 
Yuyang~Du, 
Qun~Yang, 
Soung~Chang~Liew,~\IEEEmembership{Fellow,~IEEE}
\thanks{$^*$S. Xu and L. Zhang contribute equally to this work.}
\thanks{S. Xu, L. Zhang, Y. Du, Q. Yang and S. C. Liew are with the Department of Information Engineering, The Chinese University of Hong Kong (e-mail: \{xs024, yuydu, yq020, soung\}@ie.cuhk.edu.hk, lihaozhang@cuhk.edu.hk). S. C. Liew is the corresponding author.}
\thanks{The work was partially supported by the Shen Zhen-Hong Kong-Macao technical program (Type C) under Grant No. SGDX20230821094359004.}
\vspace{-2.5em}
}

\maketitle

\begin{abstract}
Recent progress in robotics has underscored the demand for real-time control in applications such as manufacturing and healthcare systems, where the timely delivery of mission‑critical commands under heterogeneous robotic traffic is paramount for operational efficacy and safety. In these scenarios, mission‑critical traffic follows a strict deadline‑constrained communication pattern: commands must arrive within defined deadlines, otherwise late arrivals can degrade performance or destabilize control loops. In this work, we demonstrate on a real‑time software-defined radio (SDR) platform that CSMA, widely adopted in robotic communications, suffers severe degradation with contention‑induced collisions and delays disrupting the on‑time arrival of mission‑critical packets. This degradation arises under a common robotic traffic pattern where non‑critical traffic dominates the channel, while lightweight mission‑critical commands must be delivered frequently with strict deadlines over the shared medium. To address this, we propose an IEEE 802.11‑compatible hybrid TDMA/CSMA protocol that combines TDMA’s deterministic slot scheduling with CSMA’s adaptability for heterogeneous robot traffic. The protocol achieves collision-free, low-latency mission-critical command delivery and IEEE 802.11 compatibility through the synergistic integration of sub-microsecond PTP-based slot synchronization, a three-section superframe with dynamic TDMA allocation for structured and adaptable traffic management, and beacon-NAV protection to preemptively secure critical communication applications from interference. Emulation experiments on a real-time SDR testbed show that the proposed protocol reduces missed‑deadline errors by 93\% compared to the CSMA baseline under a robotic traffic setup at an overall aggregate channel load of 77.1\%, wherein 99.9\% of the traffic is from non time-critical applications and 0.1\% of the traffic is from deadline-constraint applications. In a high‑speed robot path‑tracking Robot Operating System (ROS) simulation, the protocol lowers root mean square trajectory error by up to 90\% compared with the CSMA baseline, while maintaining throughput for non‑critical traffic within ±2\%.
\end{abstract}


\section{Introduction}\label{sec-I}
Robotics has undergone remarkable advancements in recent years, playing critical roles in domains such as manufacturing \cite{calvo2021reliable}, healthcare \cite{fosch2021healthcare, chen2024llm, holland2021service}, and autonomous systems \cite{rayhan2023artificial}. Multi-robot cooperation has emerged as a key enabler for complex robotic applications that require seamless coordination among multiple devices, such as collaborative assembly \cite{keshvarparast2024collaborative}, warehouse automation \cite{dhaliwal2020rise}, and search-and-rescue missions \cite{queralta2020collaborative}. At the core of robot cooperation lies real-time control, which relies heavily on the precise and timely delivery of control commands between robots. Delays or disruptions in command transmission can degrade team performance or even precipitate severe safety hazards in high-stakes applications  \cite{nankaku2022maximum}.

As the number of robots grows rapidly in a multi-robot system, communications between robots are becoming increasingly data-intensive. The massive amount of information exchanged across different devices --  ranging from sensor data update to synchronized actuation -- shares the same communication medium, making efficient multiple access a significant challenge for the timely and reliable delivery of mission-critical control commands.

Carrier Sense Multiple Access (CSMA) is one of the most widely adopted protocols in robotic networks, known for its simplicity, flexibility, and compatibility with diverse network topologies \cite{gielis2022critical, jawhar2018networking}. This popularity is largely attributed to CSMA’s ability to facilitate medium sharing among multiple nodes without the need for intricate scheduling, as exemplified by its use in Wi-Fi networks for robotic applications \cite{jawhar2018networking, bonald2012performance}.

However, as the number of cooperating robots and their traffic loads increase, CSMA suffers from severe contentions between data streams, resulting in elevated latency when traffic congestion happens due to transmission back-offs \cite{bonald2012performance, obelovska2021performance}. In multi‑robot cooperation, contention between routine data and time‑critical commands introduces delays that compromise the responsiveness required for real‑time tasks such as path following and formation control. Moreover, multi‑robot teams are inherently dynamic: robots may join or leave, links usages fluctuate, and task priorities shift. In such environments, static or slow‑to‑adapt access schemes quickly fall out of synchronization with demand, wasting resources on inactive nodes while starving newly critical flows \cite{rizk2019cooperative}. Practical deployments must also scale to multiple robots and mixed‑traffic loads without collapsing best‑effort throughput; otherwise, safeguarding time-critical commands may undermine large‑volume data and event‑driven messages unnecessarily.

A practical robotic Medium Access Control (MAC) protocol must meet three concurrent challenges: (i) strict latency guarantees for time-sensitive command traffic under variable contention, (ii) rapid reconfiguration under dynamic topology and heterogeneous loads, and (iii) scalability in mixed‑traffic intensity without collapsing non‑time-critical long-term throughput.

Prior robotic MAC protocols largely relied on either pure CSMA or pure TDMA, yet both are inadequate for real-time multi‑robot workloads. In \cite{piguet2010mac}, the authors adopted the non‑beacon enabled IEEE 802.15.4 CSMA/CA mode to keep nodes continuously active and deliver broadcasts immediately without waiting for coordinator beacons or slot schedules. However, CSMA may suffer from backoffs, hidden terminals, and collision bursts as number of communicating robots increases, leading to unbounded latency and jitter that threaten real‑time deadlines. Pure TDMA‑based designs, such as UMCI‑MAC \cite{centelles2020underwater} for underwater cooperation and RoboMAC \cite{patti2020novel} for low‑rate mobile robotic teams, avoid collisions and bound delay through deterministic slot scheduling, but have their own limitations. Their rigid cycle structure makes it difficult to support heterogeneous and dynamic robot traffic: urgent or bursty flows must often wait for the next cycle, while best‑effort traffic is inefficiently handled. There is a need for a specially designed communication protocol that ensures deterministic delivery of time-sensitive traffic while accommodating the diverse and dynamic demands of latency-insensitive traffic in multi-robot systems.

Researchers have explored hybrid TDMA/CSMA protocols that combine 1) deterministic scheduling in TDMA with 2) flexibility of random access in CSMA. However, these designs were developed for generic wireless networks or other applications rather than robotic applications. They typically assume static or slowly varying conditions, prioritizing average throughput while overlooking strict deadlines. These limitations make them unsuitable for hard‑deadline command loops in multi‑robot cooperation over shared mediums.

The hybrid TDMA/CSMA protocol in \cite{wang2018hybrid} targeted a fixed single‑hop star with long coordinator‑scheduled superframes, where TDMA handshakes and “slot borrowing” only take effect in the next superframe -- leading to slow reallocation and weak support for frequent join/leave or urgent preemption. In \cite{gilani2013adaptive}, the MAC splits the contention access period into TDMA/CSMA under IEEE 802.15.4 standard and achieves energy and throughput gains at low rates, but its low data rate and long backoffs are incompatible with hard‑deadline robot control loops. In  \cite{liu2013hybrid}, soft‑TDMA divides associated nodes into coarse time‑frames with intra‑group CSMA. However, due to contention and only loose frame‑level synchronization, it offers no hard delay bounds. The scheme also provides weak isolation, leaving strict command‑deadline guarantees unmet under mixed robotic traffic.

This paper proposes a TDMA/CSMA framework tailored to real-time multi-robot cooperation. Our contributions are summarized as follows.

\textbf{1. Communication Pattern Analysis}: We analyze the communication pattern of time-sensitive control traffic in practical multi-robot applications running on Robot Operating System (ROS). Based on our observations on these ROS applications, we introduce the concept of deadline-constrained communication, a special type of real-time traffic that requires all communication behaviors to be completed within a strict, pre-defined deadline. We also investigated how packet contentions in CSMA affect such communication traffic via demonstration experiments on Software-Defined Radio (SDR) platform. See Section \ref{sec-II} for details.

\textbf{2. Hybrid Protocol Design}: We develop an IEEE 802.11‑compatible hybrid TDMA/CSMA system to address packet delivery failures in multi-robot communications with the deadline-constrained setup. By integrating TDMA’s deterministic scheduling with CSMA’s flexibility, our approach leverages a high-precision Precision Time Protocol (PTP) synchronization mechanism \cite{liang2021design} and a novel frame structure design to allocate collision-free time slots for critical commands. Meanwhile, CSMA is used to maintain standard compatibility and system flexibility, aiming to handle less urgent traffic efficiently. See Section \ref{sec-III} for details.

\textbf{3. Experimental Validation}: We validate our framework through SDR-based signal transmission experiments under realistic multi-robot traffic conditions. We use the resulting data for emulation experiments in a ROS environment employing the PurePursuit path-tracking algorithm for performance assessment. 
Experimental results demonstrate that the system achieves deterministic, low-latency delivery of mission-critical commands without degrading the throughput of best-effort, non-critical traffic compared to traditional CSMA in multi-robot scenarios. See Section \ref{sec-IV} for details.

\section{Motivations}\label{sec-II}
\subsection{Data Traffic in Robotic Networks}\label{sec-II-A}

In practical robotic networks, non‑critical robot traffic often dominates throughput, while lightweight mission‑critical commands must be delivered frequently under strict deadlines. In collaborative manufacturing\cite{keshvarparast2024collaborative}, camera streams and logs occupy most bandwidth, yet small control commands such as velocity or pose updates demand tight timing guarantees. A similar pattern arises in surgical robots\cite{rizk2019cooperative}, where high‑definition video and sensor traffic prevail, but small critical commands like motion corrections or micro‑positioning updates face the strictest latency constraints to avoid risks.

\begin{figure}[htbp]
\centerline{\includegraphics[width=\linewidth]{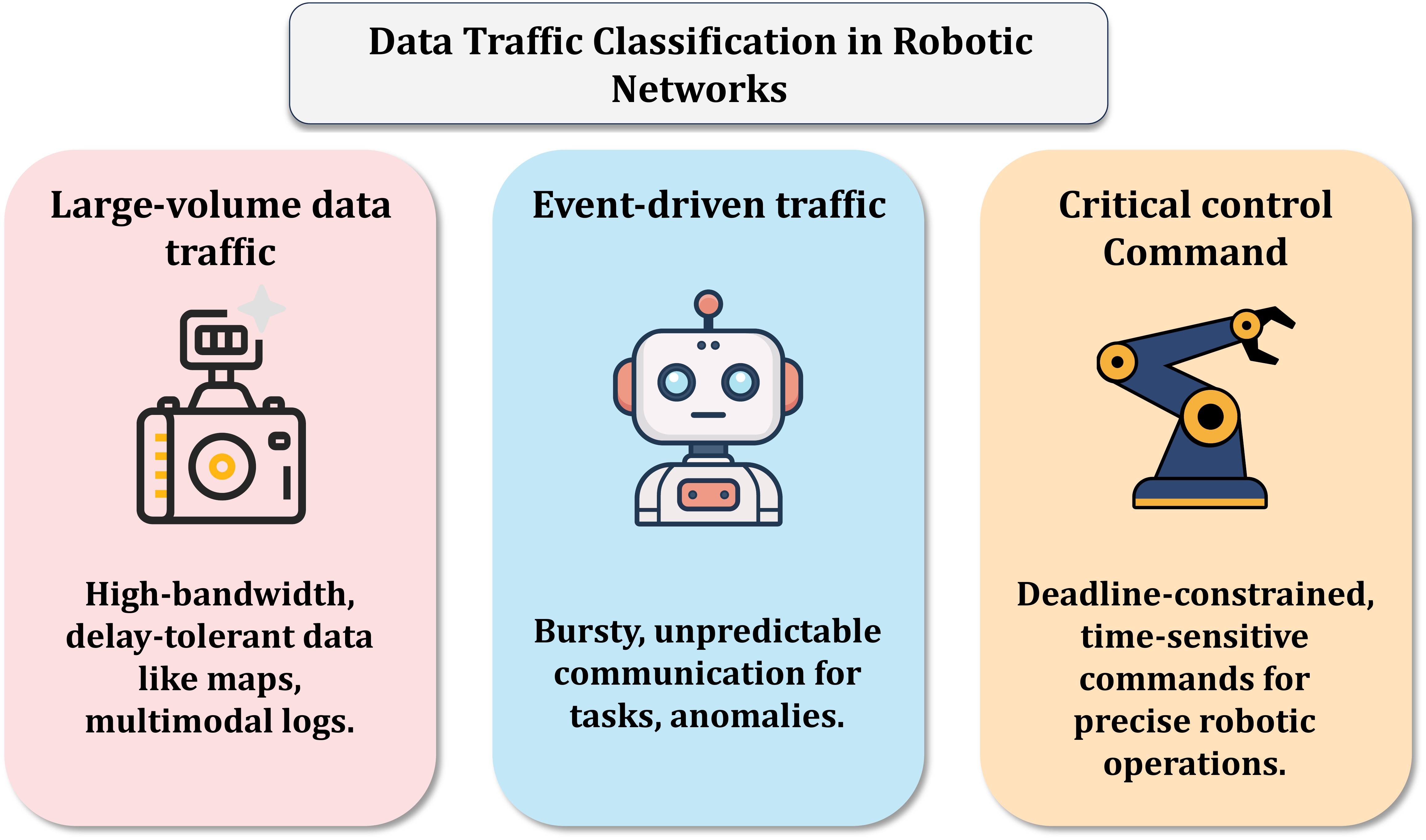}}
\centering
\captionsetup{font={small}}
\caption{ Three classes of robotic communication traffic being considered in this paper.}
\label{fig_1_robot_traffic_class}
\end{figure}

This paper classifies data traffic in a multi-robot communication network into three distinct types according to their functional roles and communication requirements: 1) large-volume data traffic, 2) bursty event-driven traffic, and 3) mission-critical control traffic. As illustrated in Fig. \ref{fig_1_robot_traffic_class}, this classification reflects the diverse operational needs of robotic systems, with each traffic class serving a unique purpose.

The first class, \textbf{large-volume data traffic}, refers to communication flows that involve substantial data exchanges among multiple robots, such as map synchronization, dataset updates, or the uploading of multimodal system logs (e.g., images, video streams).  Although these data exchanges require considerable bandwidth to maintain system consistency, they are generally less sensitive to transmission delays \cite{gutierrez2018time}. As a result, such traffic is well-suited to contention-based protocols commonly adopted in shared-medium networks, which efficiently handle bulk transfers when strict real-time guarantees are not the primary concern.

The second class, \textbf{event-driven traffic}, is characterized by bursty communication flows triggered by unpredictable events, such as task assignments from a human user to the robotic system or system anomalies detected by distributed sensors. 

For example, the integration of multimodal large language models (MLLMs) has enabled robots to understand human vocal input and generate appropriate responses to fulfill user requirements. The task-fulfillment process, as illustrated in recently proposed LLM-powered robotics and automation frameworks \cite{driess2023palm,mon2025embodied}, typically involves complex interactions between humans and multiple robots. For instance, in the LLMind framework \cite{cui2024llmind, du2025llmind}, a master robot receives human commands and sends polling messages to its slave robots to query their states. Each slave robot then reports its current status (e.g., availability, position, computational capability) back to the master for decision-making, which is followed by task decomposition at the master node and subtask execution at the slave devices. The above communication process is sporadic and bursty, given that: 1) the whole process is triggered by external, unpredictable human inputs, and 2) the response time of LLM-supported robots\footnote{Both master and slave robots in LLMind-like systems are equipped with LLM-empowered agents.}  is variable due to fluctuating task complexity and device states. The event-driven traffic in this scenario is also less sensitive to communication latency. Furthermore, the primary source of delay is the LLM inference process on robots, which may take several tens of seconds even with lightweight embedded LLMs.

The bursty event-driven traffic in the above scenarios is well-suited to CSMA-like protocols, where device contention can provide a more flexible and efficient access method than centralized scheduling.

The third class, \textbf{mission-critical control traffic}, refers to time-sensitive communication flows in multi-robot systems for control purposes. Unlike time-insensitive multi-robot tasks, such as those demonstrated with the LLMind framework \cite{cui2024llmind, du2025llmind}, in which a high-level subtask description is assigned to a device and get locally executed under the control of the embedded AI in the robot, a time-critical control command is generated by a central controller or a master robot, and it contains specific control messages that directly regulate the action of a slave robot. Typical examples of this type of traffic include the control of high-speed motors or real-time trajectory updates for distributed robots. As the control of robotic operations requires ultra-high precision, the control message must be delivered timely to ensure the operation is accurate and updated. Even microsecond-scale delays can destabilize control loops, leading to potential system failures \cite{gielis2022critical}.

To support the timely distribution of mission-critical control messages from the master robot to slave robots, ROS has adopted a Publish-Subscribe (Pub/Sub) architecture built atop the Data Distribution Service (DDS) middleware. In the Pub/Sub architecture, the master node publishes control messages (e.g., velocity commands) to designated topics, typically using the Real-Time Publish-Subscribe (RTPS) protocol over TCP. Meanwhile, communication requirements are specified via Quality of Service (QoS) parameters within DDS, among which the QoS DEADLINE policy is pivotal for real-time robot controls --- it specifies the maximum allowable interval between consecutive messages to enforce stringent timing requirements. For example, a 100 ms deadline requires the publisher to send an update at least every 100 ms, and the subscriber (e.g., an actuator controller that executes the command) expects to receive messages within this timing window. If this timing constraint is violated, probably because of delays in message generation or network transmission, the DDS middleware triggers events to alert the system. In such cases, the delayed command is no longer useful for execution, since the control action it represents has already expired.

\begin{figure}[htbp]
\centerline{\includegraphics[width=0.8\linewidth]{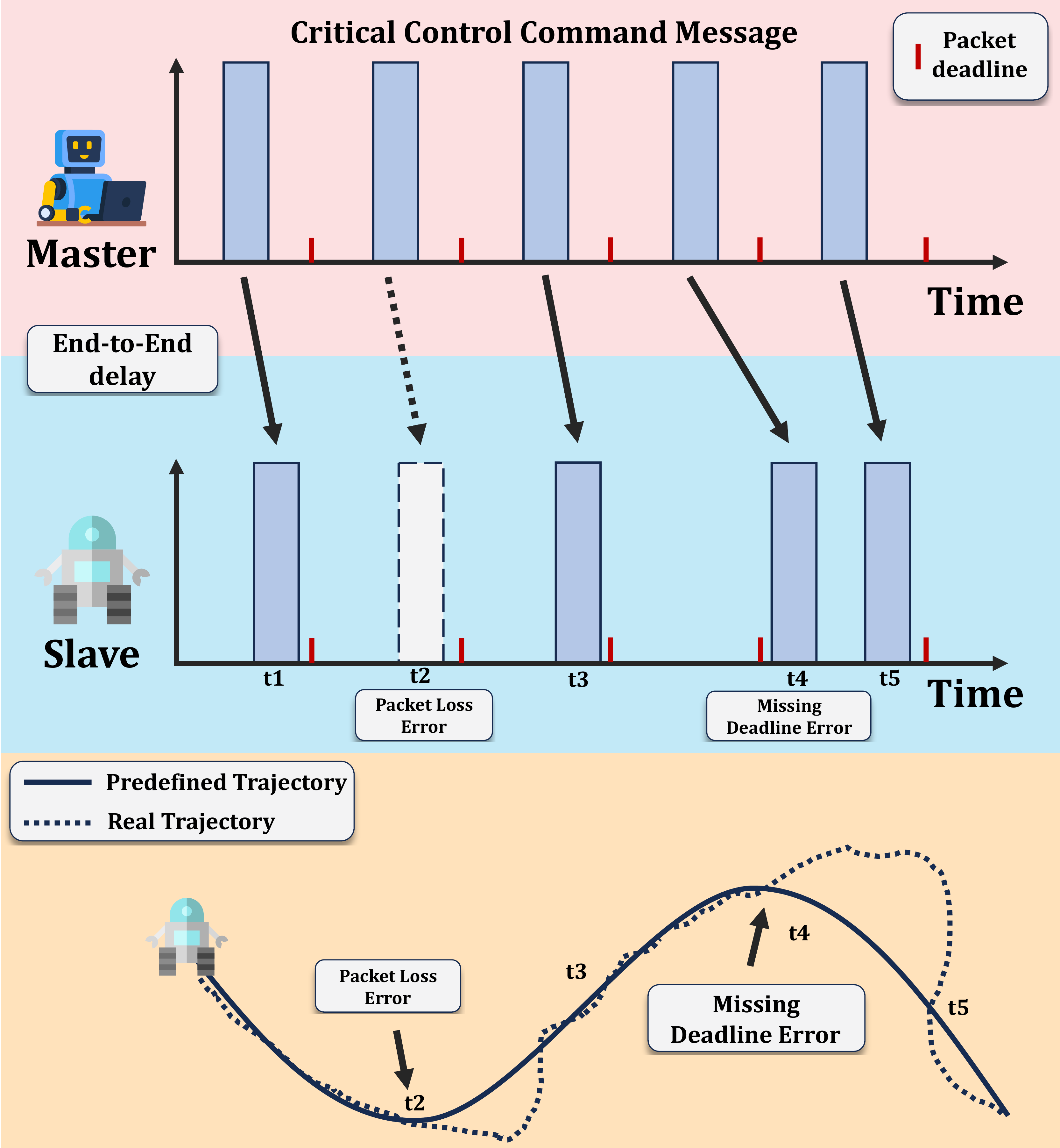}}
\centering
\captionsetup{font={small}}
\caption{Illustration of the communication deadline in a practical ROS application.}
\label{fig_mts}
\end{figure}


In the context of mission-critical control traffic, only packets arriving within their designated QoS DEADLINE window are considered valid ones with sufficient information freshness. This is illustrated in Fig. \ref{fig_mts}, which depicts a path-tracking scenario where a slave robot follows a master through an S-curve. The slave relies on periodic control packets from the master for trajectory adjustments. At time t1, the robot receives a command to adjust its steering angle for the initial turn of the S-curve. At t2, the packet is lost, and the slave robot fails to maintain a good trajectory when exiting the initial turn. At t3, the trajectory of the slave is corrected thanks to the successful delivery of packet 3. And then, due to a sudden network delay, the subsequent command, scheduled for delivery at t4, fails to arrive before its deadline. By the time it arrives, the robot’s position and orientation could differ significantly from what the control message was designed for, making the slave fail to maintain an appropriate trajectory once again. This scenario highlights the importance of timely packet delivery and the problem caused by outdated control information, explaining the necessity of the deadline constraints in the robotic communication process.

Given the above background, this paper considers excessive end-to-end delay (that may lead to missed QoS deadlines) as a form of packet transmission failure, in addition to the conventional packet transmission failure defined by the loss of a packet. Here, the end-to-end delay denotes the total time elapsed from the moment a command is issued by a master node to its successful delivery at a slave node, which includes the time consumed in signal propagation, buffer queuing, protocol stack processing, and network transmission. 

We emphasize that conventional CSMA protocols, such as CSMA/CA in Wi-Fi networks, are inadequate for such deadline-constrained traffic. That is, the contention-based multiple access mechanism in CSMA results in packet collisions and backoff delays under high traffic loads, leading to long and unpredictable queuing latency in the transmitter’s buffer. When diverse traffic types in a multi-robot system compete for limited bandwidth, CSMA may fail to prioritize time-sensitive control commands, resulting in packet transmission failures due to missed deadlines (see Section \ref{sec-II-B} below for further discussions).

\subsection{Motivation Demonstrated via Emulation Experiments}\label{sec-II-B}

This subsection evaluates the performance of the conventional CSMA protocol for mission-critical control traffic in a practical multi-connection communication setting. The experimental results motivate the introduction of TDMA into the system. 

\begin{figure}[htbp]
\centerline{\includegraphics[width=0.8\linewidth]{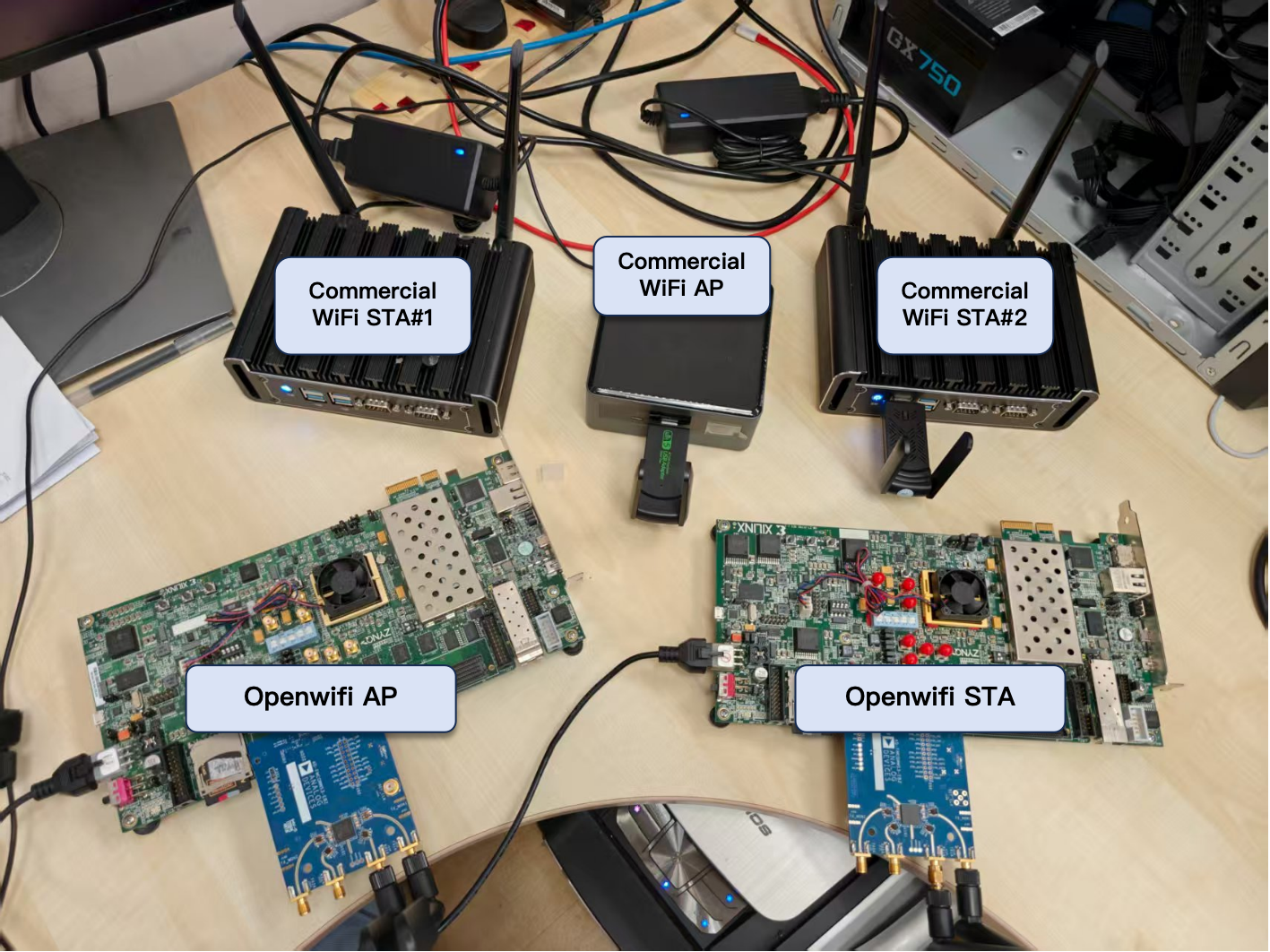}}
\centering
\captionsetup{font={small}}
\caption{A photo of the real-world system setup}
\label{fig_real_demo}
\end{figure}

To monitor the dynamic behavior of practical random access, especially those related to the contention and back-off of a packet, we built our real-time testbed upon Openwifi  \cite{jiao2020openwifi}, an open-source IEEE 802.11 Wi-Fi system on SDR platforms. Thanks to the high-speed field-programmable gate array (FPGA) integrated into the Xilinx Zynq-7000 System-on-Chip (SoC), the testbed supports real-time signal processing and network operations. Fig. \ref{fig_real_demo} shows the real-world system setup, in which a point-to-point communication was established with two Xilinx ZC706 boards, each equipped with an AD9361 RF board. One board serves as the master (server), while the other serves as the slave (client). Additionally, a commercial Wi-Fi access point (AP) and two commercial Wi-Fi stations (STAs) are configured to emulate diverse traffic over the air. All these devices are set to the same 2.4GHz WiFi channel (specifically, Band 3 of the IEEE 802.11n standard) to ensure the existence of traffic contention.

\begin{figure}[htbp]
\centerline{\includegraphics[width=\linewidth]{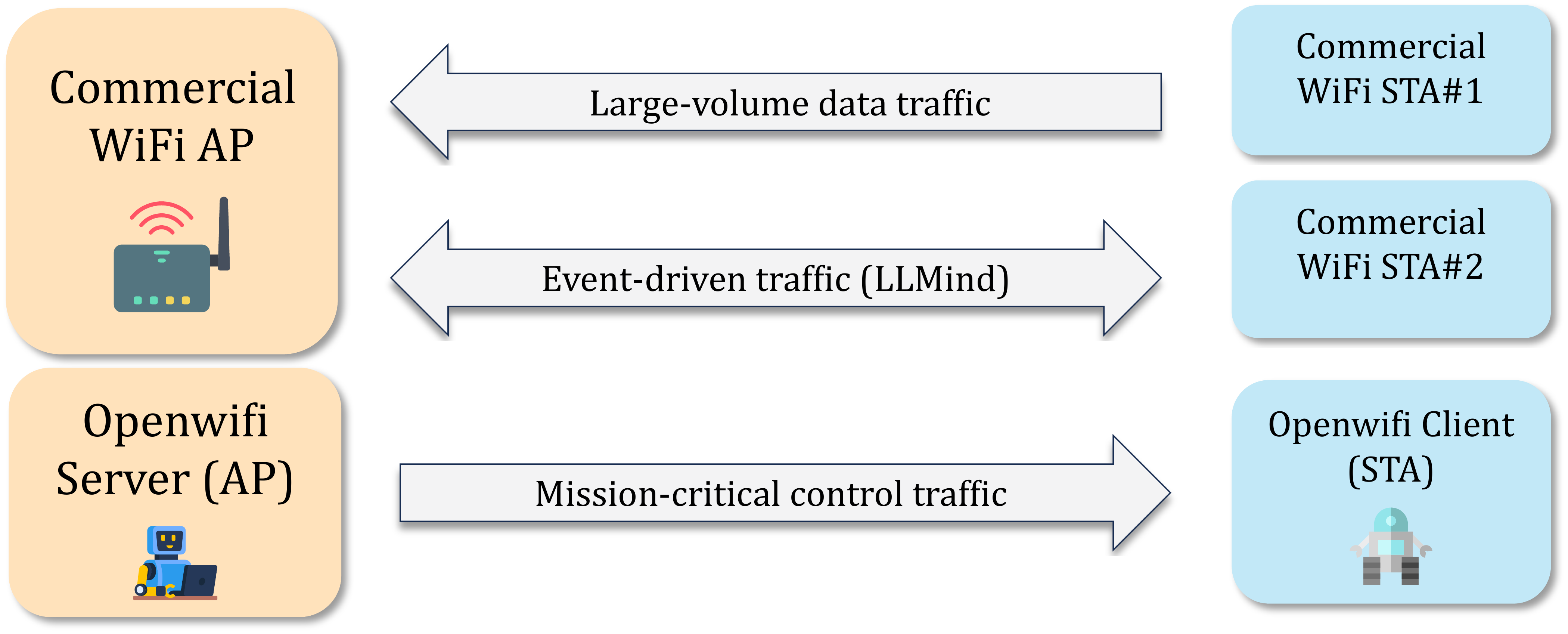}}
\centering
\captionsetup{font={small}}
\caption{Multi robot traffic scenario}
\label{fig_exp_setup}
\end{figure}

As a supplement of Fig. \ref{fig_real_demo}, Fig. \ref{fig_exp_setup} and Table \ref{tab:real_time_scenario} give further details on how the three types of traffic flow mentioned above were realized in our multi-connection scenario:

1) The large-volume traffic was represented by the periodic video transmissions between the commercial AP and commercial STA \#1. This is to simulate the view sharing mechanism between robots.

2) The bursty event-driven traffic was represented by the coordinator-agent communication in an LLMind system, which is realized with the link between the commercial AP and commercial STA \#2. To simulate the randomness in the transmission process, we let the transmission period follow a Poisson distribution with a mean inter-arrival time of 40 seconds as an emulation of human behavior. Within each cycle, the server response time follows an exponential distribution with a mean of 25 seconds to simulate the random response latency of LLM agents.

3) The mission-critical robot command traffic was represented by a deadline-constrained packet in ROS, with the QoS deadline request set to 100ms. That communication link was realized by the Openwifi AP and the Openwifi STA\footnote{ The mission-critical robot command traffic can also be simulated with commercial AP and STAs. Here we use the more complex openwifi platform to realize that traffic link to provide a fair comparison for our later experiment: our later implementation of the hybrid TDMA/CSMA system was built upon the openwifi system, simulating the pure CSMA case with the same hardware platform would provide a fair baseline for later comparisons.}. 

\begin{table}[htbp]
\caption{A Case for Real-Time Multi-Connection Scenario}
\centering
\small
\begin{threeparttable}
\begin{tabular}{|>{\centering\arraybackslash}p{2.2cm}|
                >{\raggedright\arraybackslash}p{\dimexpr\linewidth-2.2cm-4\tabcolsep-2\arrayrulewidth}|}
\hline
\textbf{Traffic Type\tnote{3}} & \textbf{Description} \\
\hline
large-volume data traffic & Periodic transmission from Client to Server: 5 MB data traffic every 3 seconds \\
\hline
\makecell[c]{Event-driven \\ traffic} & 
\makecell[l]{
An LLMind system with the following setup:\\
1. Server to Client: 2 KB (Polling) \\
2.Client to Server: 512 KB (Historical Data)\tnote{4} \\
3. Server to Client: 5 KB (FSM Data)\tnote{5}
} \\
\hline
mission-critical control traffic & Server to Client: 25 bytes every 100 ms via TCP with 100ms ROS QoS DEADLINE \\
\hline
\end{tabular}
\begin{tablenotes}
\scriptsize
\item[3] As mentioned in Section \ref{sec-II-A}, we consider a robotic traffic pattern at an overall aggregate channel load of 77.1\%, wherein 99.9\% of the traffic is from non time-critical applications (large-data traffic and event-driven traffic) and 0.1\% of the traffic is from deadline-constraint applications.
\item[4] Historical Data refers to previously stored interaction records and validated control scripts uploaded by the client for fast retrieval and reuse in LLMind system.
\item[5] FSM Data is the finite-state machine representation of decomposed tasks generated by the server to guide subsequent control script execution in LLMind system.
\end{tablenotes}
\end{threeparttable}
\label{tab:real_time_scenario}
\end{table}

Building on the discussion in Section \ref{sec-II-A}, the transmission of mission-critical control traffic can be categorized three states: 1) \textit{Transmission Success}, where the packet is delivered successfully within the specified QoS deadline; 2) \textit{Transmission Failure - Missed Deadline}, where the packet arrives at the client but exceeds the deadline, making it invalid due to outdated control information; and 3) \textit{Transmission Failure - Packet Loss}, where the packet is lost due to network conditions or contention. 

To clearly distinguish different transmission outcomes, we define the classification under the deadline constrained setup. Each packet is allowed one initial transmission plus up to seven retransmissions at the Openwifi MAC layer. If a packet is delivered successfully within the 100 ms QoS deadline during its initial transmission or any of the seven retransmission attempts, it is considered \textit{Transmission Success}. If a delivery eventually succeeds but its last attempt finishes after the 100 ms deadline, it is classified as a \textit{Transmission Failure – Missed Deadline}.  If all eight attempts fail to deliver the packet, the outcome is classified as a \textit{Transmission Failure – Packet Loss}. In addition, the packet loss category covers both collision induced losses and wireless channel errors, as both were considered in our SDR based experiments.

\begin{figure}[htbp]
\centerline{\includegraphics[width=0.9\linewidth]{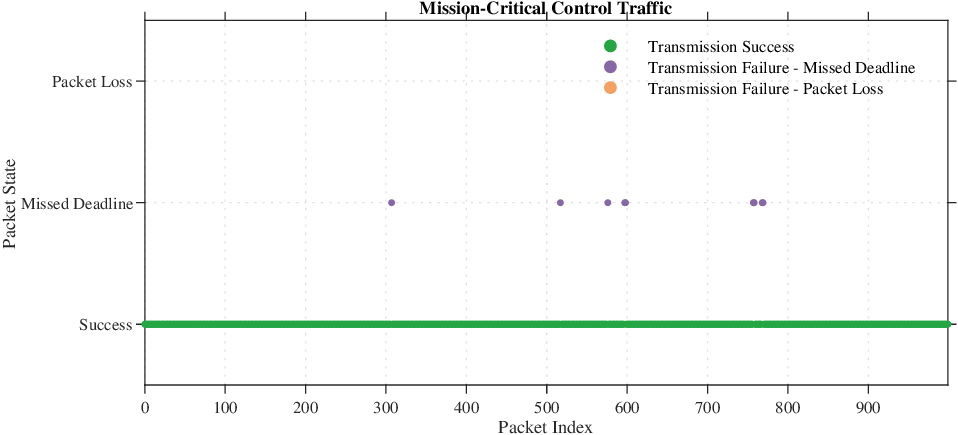}}
\centering
\captionsetup{font={small}}
\caption{Baseline experimental result of a pure CSMA system under a low‑load scenario with only one mission‑critical control traffic flow.}
\label{fig_csma_sequence_ideal}
\end{figure}

We first evaluated a low‑load scenario with only one mission‑critical control traffic flow as a baseline. The result, shown in Fig. \ref{fig_csma_sequence_ideal}, indicates that transmission failures are extremely rare, with only negligible deadline misses or packet losses observed during the test period. These results suggest that CSMA can provide acceptable reliability under light, single‑flow traffic conditions; however, this baseline case does not capture the challenges that arise under higher contention.

\begin{figure}[htbp]
\centerline{\includegraphics[width=\linewidth]{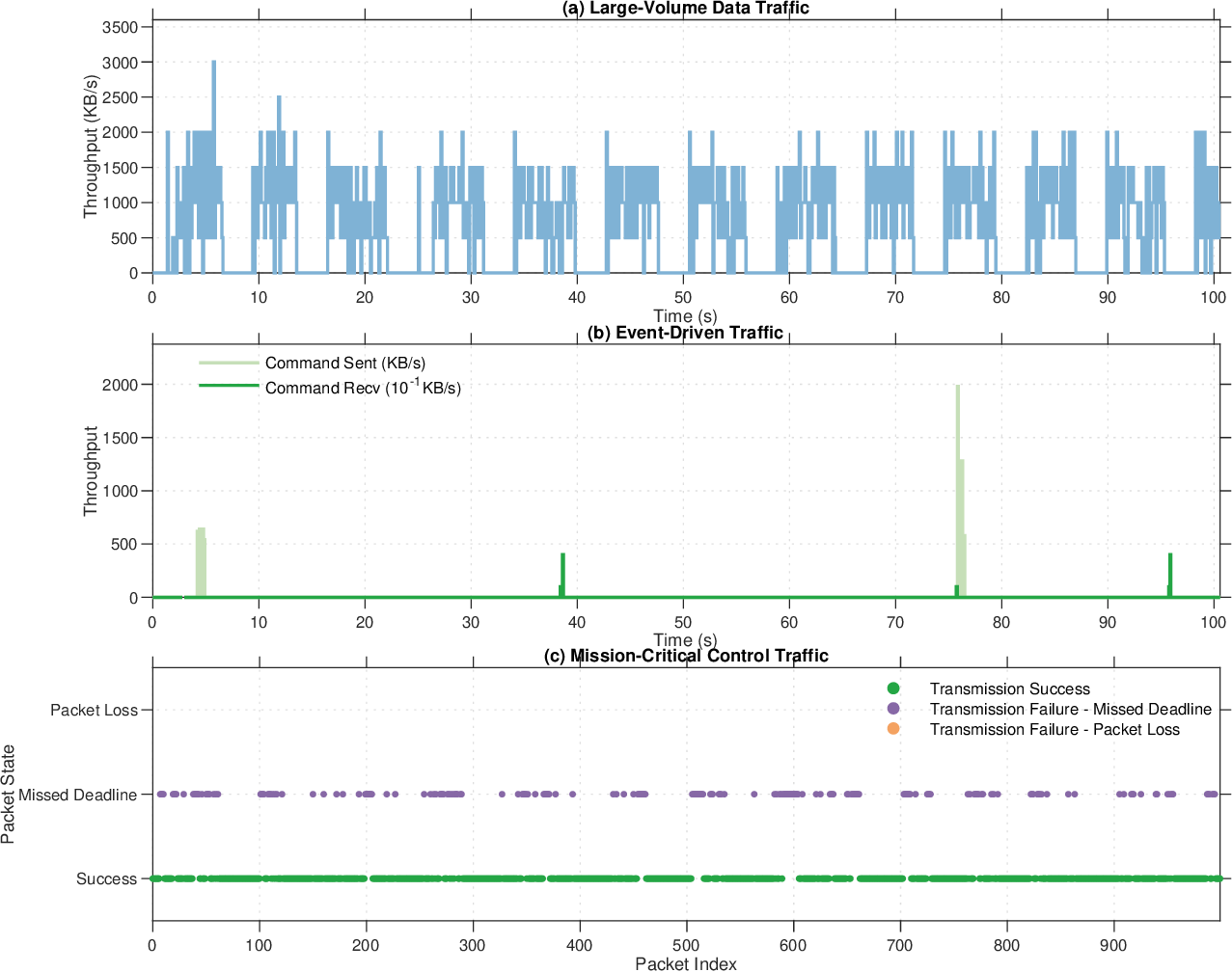}}
\centering
\captionsetup{font={small}}
\caption{The experimental result obtained in a CSMA system, which motivates our hybrid TDMA/CSMA setup. Figs.(a) and (b) give the throughput of the large-volume data traffic and the event-driven traffic, respectively,  where throughput is computed as the number of successfully delivered bytes per averaging window of 100ms. Fig. (c) shows the packet transmission state of the mission-critical control traffic in the experiment.}
\label{fig_csma_sequence_multi}
\end{figure}

Building on the multi‑connection setup described in Fig. \ref{fig_exp_setup} and Table \ref{tab:real_time_scenario}, we next evaluate the performance of CSMA when these three types of traffic flows coexist. Fig. \ref{fig_csma_sequence_multi} presents the experimental results obtained in a pure CSMA setup under this typical multi‑robot traffic scenario. In the over-1000-second test period, we found that 99.71\% of transmission failures in Fig. \ref{fig_csma_sequence_multi}(c) were caused by missed deadline, while only 0.29\% of transmission failures could be attributed to packet loss. This observation reveals a fact that the transmission uncertainty in CSMA leads to unpredictable latency and jitter, making the end-to-end delays frequently exceeding the stringent QoS deadline when back-off happens. 

Based on the evaluation result, we conclude that CSMA’s contention-based access mechanism fails to fulfill the deterministic delivery and the stringent latency requirement for deadline-constrained communication pattern in robotic control, given that the co-existence of other types of communication traffic introduces network congestion and causes mission-critical control traffic packet collisions in CSMA. These deficiencies highlight the need for a specialized protocol to ensure deterministic, low-latency delivery of critical robot commands while not affecting other types of robotic traffic.

\section{IEEE 802.11-Compatible Hybrid TDMA/CSMA Protocol for Multi-Robot Cooperations}\label{sec-III}
Given the limitation of existing IEEE 802.11 protocols in handling diverse, dynamic robotic traffic, we propose an IEEE 802.11-compatible hybrid TDMA/CSMA protocol tailored for multi-robot cooperation. Subsection A outlines the key features and design principles of the proposed protocol. Subsection B presents an overview of core architectural components within this protocol. Subsection C gives a comprehensive description of the protocol’s implementation. 

\subsection{Design Principles of the Hybrid TDMA/CSMA Protocol}

To address the limitations of existing protocols in multi-robot systems, particularly their vulnerability to packet collisions and unpredictable latency under heavy robotic traffic, we introduce TDMA for deterministic delivery of mission-critical control traffic. TDMA allocates specific time slots to each node for collision-free communication, with only one node transmitting at a time. However, meeting the required time precision of TDMA in a dynamic multi-robot environment presents significant challenges.

The first challenge in realizing TDMA within the studied multi-robot systems is achieving precise time synchronization among master and slave robot nodes. In TDMA, each node is assigned specific time slots for collision-free communication. However, in practical deployments, precise clock alignment between master and slave nodes cannot be fully guaranteed due to factors such as oscillator drift \cite{lasassmeh2010time, du2022ser, du2025reliable}. A microsecond-level timing misalignment (e.g., a maximum of 9µs is allowed in 802.11a/g/n) can cause nodes to transmit outside their assigned time slots, leading to overlapping transmissions and packet collisions between multiple nodes. 

Another challenge in implementing TDMA is realizing flexible time-slot assignment among different robot nodes. Multi-robot cooperation involves continuously changing traffic patterns driven by task variations and heterogeneous application requirements. For instance, a robot performing high-precision assembly may require more frequent control updates than one conducting routine patrol. Fixed TDMA allocations cannot efficiently adapt to such varying demands, leading to bandwidth waste or performance degradation when slot allocation does not match demand. In such cases, dynamic adjustment of TDMA slot allocations is needed to meet varying communication needs, while coordinating robots and avoiding excessive synchronization overhead or message collisions.

Third, for multi-robot cooperation, potential collisions may occur between TDMA and CSMA sections due to their different access mechanisms. CSMA’s contention‑based channel access may overlap with TDMA slots or the control slot of TDMA, causing interference under diverse traffic conditions. Therefore, an effective mechanism is required to guarantee the deterministic delivery of TDMA packets while efficiently using residual bandwidth for CSMA to manage dynamic, non-critical robot traffic. 

To address these challenges, our hybrid TDMA/CSMA protocol has three key features. First, we integrate the Precision Time Protocol (PTP) for sub-microsecond synchronization across nodes to support IEEE 802.11-compatible precise alignment of TDMA slots despite robot mobility. Second, we introduce a three-section superframe that separates traffic into: (i) deterministic TDMA slots for mission-critical control traffic; (ii) a TDMA control section for transmitting dynamic slot allocation information, and (iii) a general-purpose CSMA section for non-critical robot traffic. Finally, we employ the Network Allocation Vector (NAV) mechanism, embedded in beacon frames, to protect both the TDMA section and TDMA control information by preemptively reserving the channel. This prevents interference from opportunistic CSMA accesses and guarantees the deterministic, low-latency communication essential for multi-robot cooperation.

\begin{figure}[htbp]
\centerline{\includegraphics[width=\linewidth]{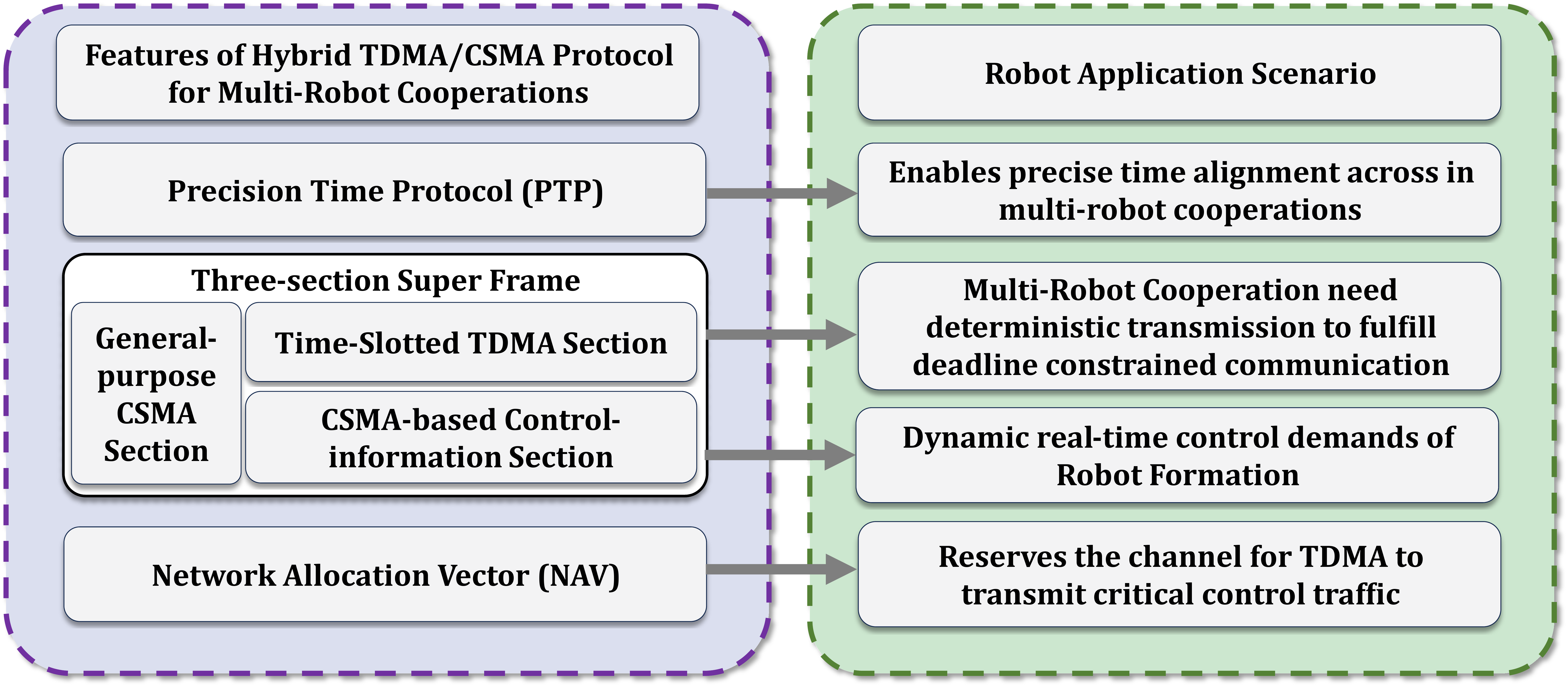}}
\centering
\captionsetup{font={small}}
\caption{Diagram of Hybrid TDMA/CSMA Protocol Features Supporting Multi-Robot Cooperation}
\label{fig_diagram_feature}
\end{figure}

Fig. \ref{fig_diagram_feature} shows the mapping between novel features developed in the hybrid protocol and the practical challenges these features aim to address. A detailed explanation of these features is given in Subsection B below.

\subsection{Key Features of the Hybrid TDMA/CSMA Protocol}

\subsubsection{Feature 1: PTP Synchronization Mechanism}\label{sec-ptp}

Precise time synchronization is essential for aligning time slots in the TDMA section. We achieve a sub-microsecond clock alignment across nodes through a three-way handshake mechanism, compensating for clock offsets and propagation delays.

In our system, a master server node acts as a synchronization coordinator as well as the AP in the IEEE 802.11 network, broadcasting beacons to slave client nodes. Each beacon, sent at the start of a superframe, carries the server’s transmission timestamp  $s_{ap,beacon}^l$ for the beacon of the $l$-th superframe. In other words, the beacon in our system serves as one of the PTP synchronization messages. Upon receiving the beacon, the $i$-th client records its arrival time, $\tilde{s}_{i,beacon}^l$, which is offset by the propagation delay $d_{i,delay}$ and clock offset $o_i$ \cite{ieee2009ieee}:

\begin{equation}
\tilde{s}_{i,beacon}^{l}=s_{ap,beacon}^{l}+{{d}_{i,delay}}+{{o}_{i}}
\end{equation}

For synchronization purposes, any one of the slave clients should respond to AP with another synchronization message (referred to as a synchronization response message in the rest of this paper) at any time during superframe section 2. We require each client accepted into the network to perform the synchronization procedure periodically. Specifically, for the synchronization process, the client embeds its transmission time, $s_{i,response}^l$, in the packet’s payload during the l-th superframe, representing the time the packet is sent according to the client’s local clock. The time $s_{i,response}^l$ is determined by the client’s local scheduling, initially based on the beacon’s arrival time, and progressively adjusted as the synchronization process continues (see Sections IV-A3 and IV-C in \cite{liang2021design}). The server records the packet’s arrival time, $t_i^l$, and includes it in the next beacon, which is received by the client. Using these timestamps and assuming symmetric propagation delays ($d_i=d_{i,0}$), clients estimate:

\begin{equation}
{{d}_{i}}=\frac{1}{2}(\tilde{s}_{i,beacon}^{l}+t_{i}^{l}-s_{ap,beacon}^{l}-s_{i,response}^{l})    
\end{equation}

\begin{equation}
{{o}_{i}}=\frac{1}{2}(\tilde{s}_{i,beacon}^{l}-t_{i}^{l}-s_{ap,beacon}^{l}+s_{i,response}^{l})    
\end{equation}

To align packet arrivals at the server, the $i$-th clients schedules its transmission time for the $j$-th slot of the $l$-th superframe section 1 as follows:

\begin{equation}
    s_{i,j}^{l}=s_{i}^{l}+{{o}_{i}}+(j-1){{T}_{s}}-2{{d}_{i}}
    \label{eqni}
\end{equation}

where $s_i^l$ is the start time of the $l$-th frame according to the counter of the $i$-th client (i.e., slot 1 begins at this time) and $T_s$ is the duration of a slot. 

\subsubsection{Feature 2: Multi-Section Superframe with Flexible Time Slot Allocation}

\begin{figure}[htbp]
\centerline{\includegraphics[width=\linewidth]{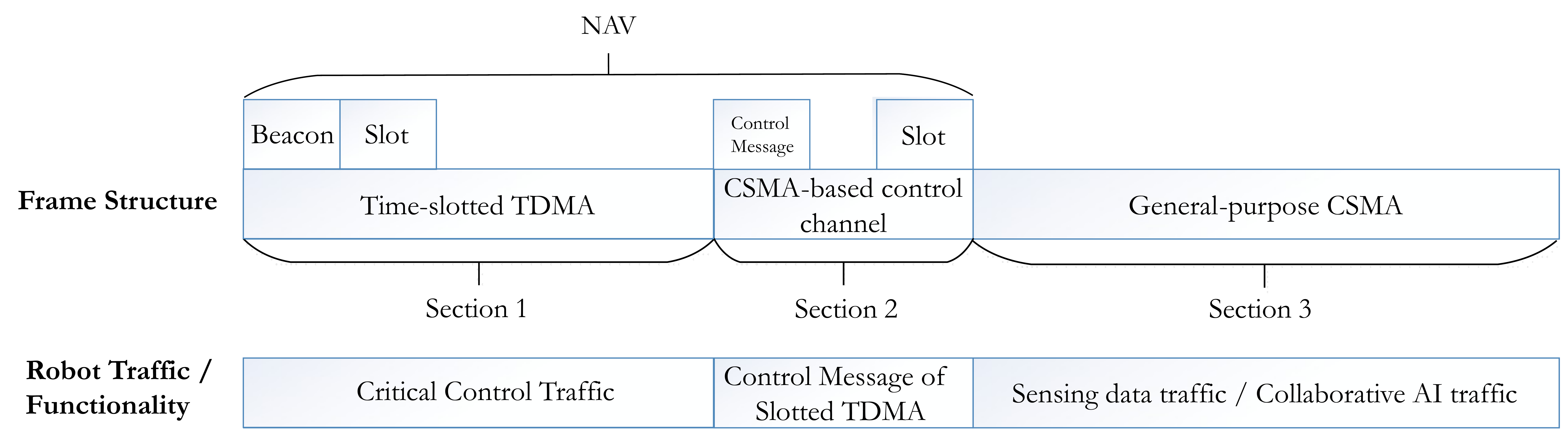}}
\centering
\captionsetup{font={small}}
\caption{Superframe Structure of the IEEE 802.11-Compatible Hybrid TDMA/CSMA Protocol for Real-Time Robot Control}
\label{fig_superframe_struc}
\end{figure}

Fig. \ref{fig_superframe_struc} shows the superframe structure applied in our hybrid TDMA/CSMA protocol. The total duration of a superframe, $T_f$, is defined as:

\begin{equation}
T_f=T_{TDMA}+T_{CSMA}^{ctl}+T_{CSMA}^{gen}    
\end{equation}

where $T_{TDMA}$, $T_{CSMA}^{ctl}$ and $T_{CSMA}^{gen}$ denote the durations of the time-slotted TDMA section, the CSMA-based control section, and a general-purpose CSMA section, respectively.
 
\textbf{\textit{Superframe Section 1 (Time-slotted TDMA Section)}}: The time-slotted TDMA section is exclusively reserved for the mission-critical traffic. Each participating node, referred to as an STA in IEEE 802.11 terminology, is assigned dedicated time slots within the superframes. The duration of the TDMA section is given by:

\begin{equation}
T_{ TDMA } =N_{ TDMA } \times \tau_{TDMA }    
\end{equation}

where $N_{TDMA}$ is the number of TDMA slots, and $\tau_{TDMA}$ is the duration of each slot, including the transmission time for critical packets and associated overhead. Each slot is allocated to a specific user through a time slot allocation described in the CSMA-based control section. 

\textbf{\textit{Superframe Section 2 (CSMA-based Control Section)}}: The control section is dedicated to transmitting control and management messages for protocol coordination and network operations. Specifically, three messages are included: 1) IEEE 802.11 standard management messages, including association, authentication, and the integration of new STAs into the network; 2) PTP synchronization messages \cite{liang2021design} (discussed in Feature 1); and 3) time-slot allocation directives. The TDMA server dynamically allocates the number ($N_{TDMA}$) and duration ($\tau_{TDMA}$) of slots for each client node based on real-time network conditions and robot application requirements. The TDMA server updates these assignments and clients configure schedules accordingly through the CSMA-based control section.

\textbf{\textit{Superframe Section 3 (General-purpose CSMA Section)}}: The general-purpose CSMA section is used for non-time-critical traffic, such as large-volume data traffic or event-driven traffic, which do not require stringent latency guarantees. Nodes access the medium via CSMA/CA, using IEEE 802.11 mechanisms for flexible, best-effort delivery.

\subsubsection{Feature 3: Beacon’s NAV protection}

To protect the TDMA section and the control section from interference, we use the NAV of the beacon frame to reserve the medium, i.e., prevent non-relevant STAs from initiating transmissions. The length of the beacon NAV is set as the exact duration of the TDMA section plus the CSMA-based control section:
\begin{equation}
T_{NAV}=T_{TDMA}+T^{ctl}_{CSMA}    
\end{equation}

And the beacon frame, compliant with the IEEE 802.11 standard, is broadcast by the server node at the beginning of each superframe. 

\subsection{Hybrid TDMA/CSMA Protocol Implementation}

\begin{figure*}[htbp]
\centering
\captionsetup{font={small}}
\includegraphics[width=\linewidth]{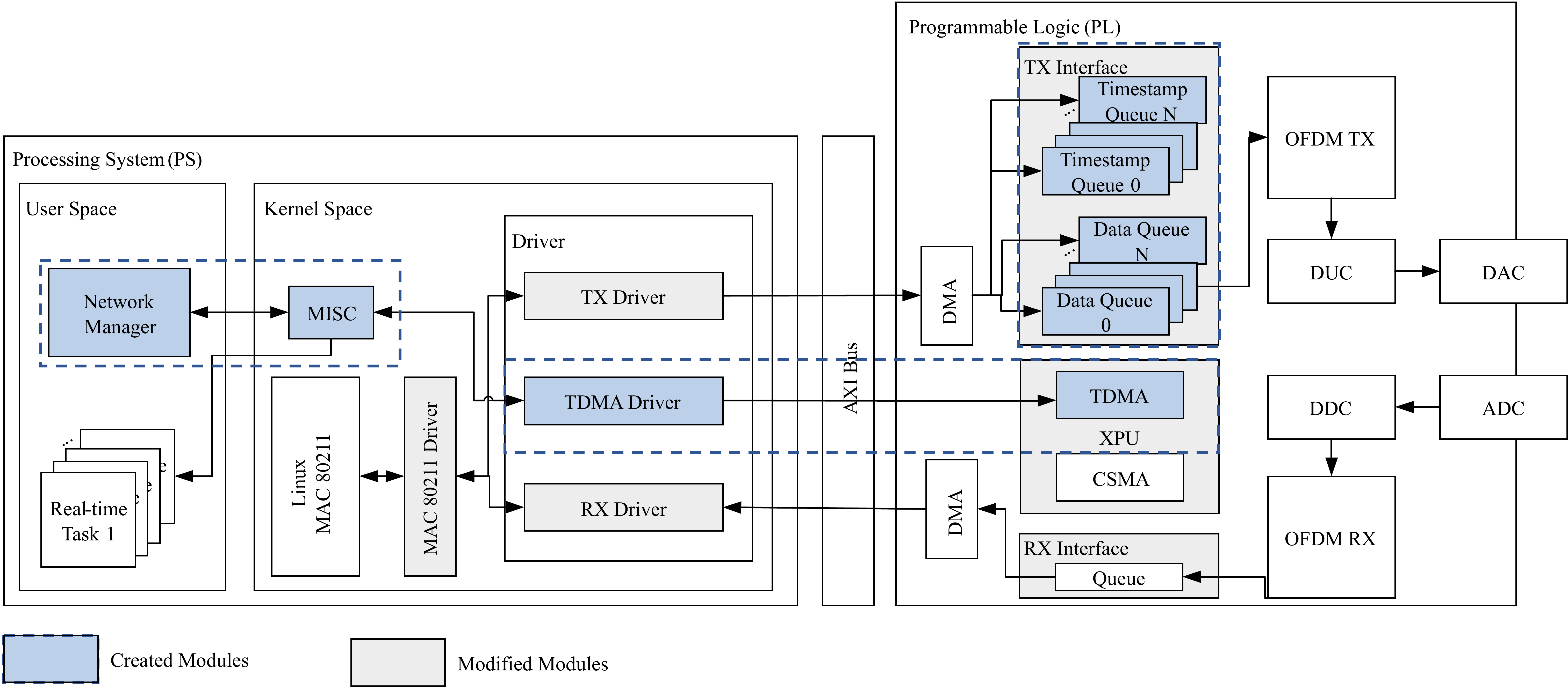}
\caption{Implementation Architecture of Hybrid TDMA/CSMA Protocol, Featuring: (1) Dual-Queue System for Traffic Prioritization, (2) NAV Protection and PTP-Based Synchronization, and (3) Low-Latency Middleware and Dynamic APIs}
\label{fig_imple_struc}
\end{figure*}

Our implementation is based on the Openwifi project \cite{jiao2020openwifi}, as shown in Fig. \ref{fig_imple_struc}. This project divides IEEE 802.11 system functionality into hardware and software domains. The hardware domain, implemented on the FPGA, handles time-critical PHY and MAC functions, including modulation and CSMA/CA-based access control, with Verilog-implemented blocks for carrier sensing. The Analog Devices AD9361 RF frontend supports wireless communication with precise timing at a 61.44 MHz sampling rate. The Advanced eXtensible Interface (AXI) Bus connects software and hardware, providing a 600 MB/s channel for packet data and control signals. The software domain, hosted on the ARM processor running Linux, includes the Openwifi driver and mac80211 subsystem for protocol scheduling and network configuration. On the hardware side, we extend Openwifi project by adding three features into the MAC layer, namely 1) PTP synchronization, 2) packet timestamping, and 3) multi-queuing. On the software side, we develop a TDMA driver with a network manager API to realize packet transmission scheduling and TDMA slot allocations. In the following, we will discuss these novel implementations respectively.

\subsubsection{PTP Synchronization implementation}
In our system, the PTP synchronization messages (beacon and synchronization response message as discussed in Section \ref{sec-ptp}), exchanged exclusively during the CSMA-based control section, carry 64-bit nanosecond-resolution timestamps. At reception of the $i$-th client, the hardware logic updates local clock offsets $o_i$ and propagation delay $d_i$ estimates, which are then applied to adjust transmission schedules $s_{i,j}^l$ for its subsequent TDMA slots. Specifically, $o_i$ and $d_i$ are returned to the TDMA driver for the calculation of $s_{i,j}^l$ as in Eq. \ref{eqni}. 

\subsubsection{Packet Timestamping and multi-queue mechanism}
To support deterministic TDMA scheduling, we implement a packet timestamping mechanism. That is, the TDMA driver of the $i$-th client computes the transmission timestamp $s_{i,j}^l$ for the TDMA packet that will be transmitted at $j$-th slot of the $l$-th superframe. When a TDMA packet is generated from the application layer for transmission, the TDMA driver writes the packet data to a hardware data queue and its corresponding timestamp to a dedicated timestamp queue. At the hardware level, the FPGA maintains a high-resolution counter operating at the system clock frequency of 61.44 MHz to support nanosecond-level timing. The TDMA XP block continuously monitors the head of the timestamp queue and compares it with the current counter value. When the counter matches the queued timestamp, the hardware immediately triggers packet transmission by extracting the corresponding packet from the data queue and initiating the PHY-layer processing pipeline.

Furthermore, we use a multi-queue architecture to separate TDMA and CSMA traffic based on assigned port numbers at the robotic application layer. Specifically, if a packet is identified as TDMA, both the packet and its timestamp are placed into the TDMA queue and the corresponding timestamp queue. CSMA traffic is further classified into control management messages, placed in the CSMA-control queue for the control section, and general data packets, placed in the CSMA-data queue for the general-purpose section. All CSMA queues operate in a first-in-first-out manner. The multi-queue mechanism integrates directly with the protocol’s three-section superframe structure for deterministic delivery of mission-critical TDMA traffic and flexible non-critical CSMA traffic.

\subsubsection{Network Manager API}
The network manager is implemented as a user-space daemon to provide socket-based APIs for real‑time protocol configuration and control. Through these APIs, a robotic application can request adjustments to TDMA slot parameters such as the number of slots ($N_{TDMA}$), slot durations ($\tau_{TDMA}$), or reassigning slot ownership among active robots to better match changing communication loads and task priorities. When a configuration change is issued, the network manager constructs an updated allocation table, packages it into an IEEE 802.11 compliant management frame and transmits it during the CSMA based control section. 

The middleware layer provides a publish–subscribe interface with the TDMA scheduler. Robotic applications register for slot based transmission by submitting a subscription message with explicit QoS parameters such as the maximum allowable deadline, the desired update frequency (e.g., 10 Hz steering updates), and the traffic priority level. The network manager matches these QoS requirements with the current network state to generate a TDMA slot map, which is then queued in the control section for broadcast to all nodes. In kernel space, the TDMA driver uses this slot map to configure high resolution timers to trigger packet dequeue operation at assigned slot boundaries.

\section{Performance Evaluation of the hybrid TDMA/CSMA Protocol}\label{sec-IV}
\begin{figure}[htbp]
\centerline{\includegraphics[width=\linewidth]{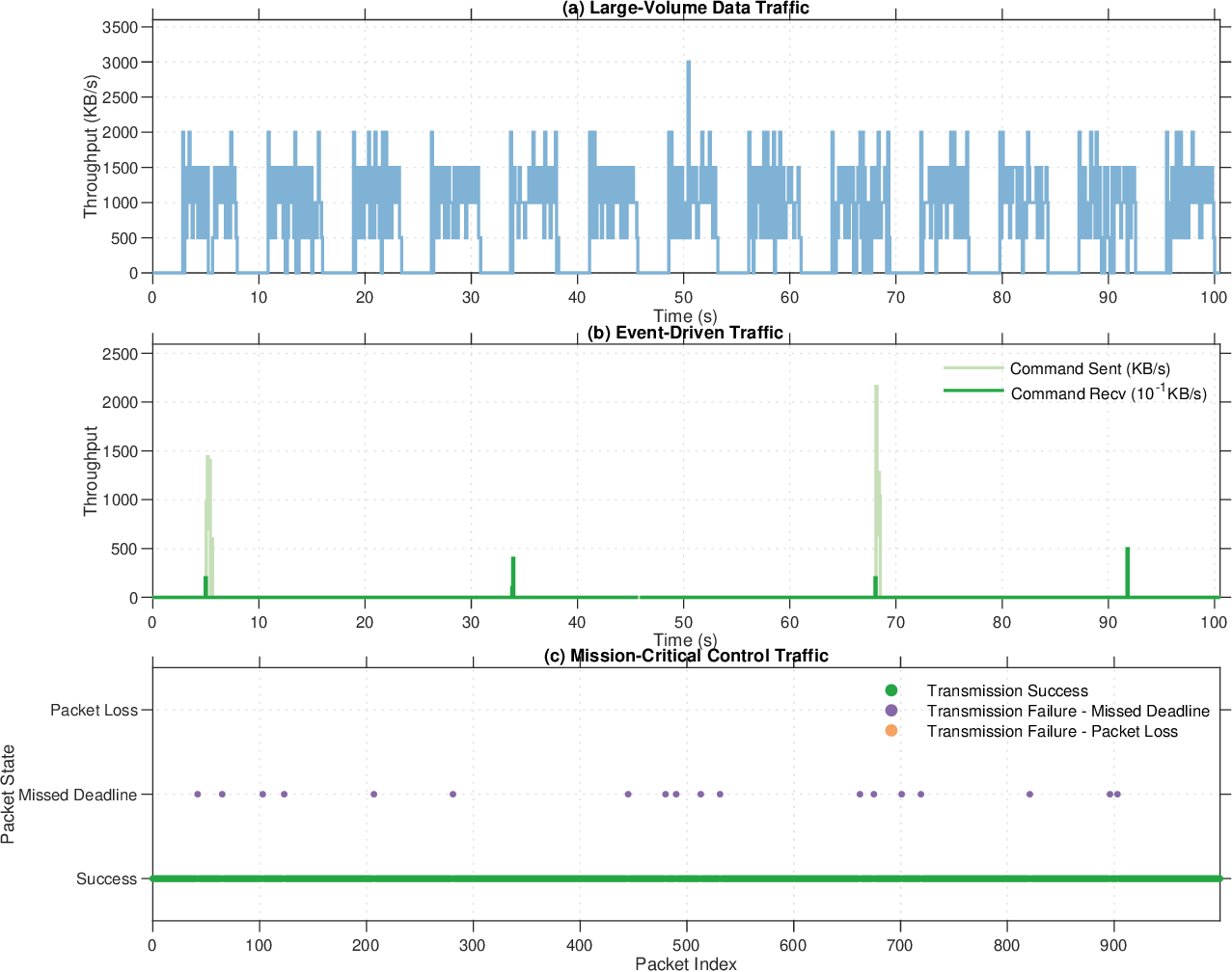}}
\centering
\captionsetup{font={small}}
\caption{The experimental result obtained from the proposed hybrid TDMA/CSMA implementation. Subfigure descriptions are the same as those in Fig. \ref{fig_csma_sequence_multi}}
\label{fig_tdma_sequence_multi}
\end{figure}

Fig. \ref{fig_tdma_sequence_multi} shows the temporal performance of the proposed hybrid TDMA/CSMA protocol under varying robotic traffic. The experimental setup is the same as in Section \ref{sec-II-B} except for replacing the CSMA protocol with the hybrid scheme for mission-critical control traffic transmission.

The comparison of non-critical traffic shown in Fig.\ref{fig_tdma_sequence_multi}(a) and \ref{fig_tdma_sequence_multi}(b) with their CSMA counterparts in Fig. \ref{fig_csma_sequence_multi} shows that the throughput of event-driven traffic and large-volume data traffic remains virtually unchanged. Specifically, the average throughput of non-critical flows fluctuates by less than 2\% between the two protocols throughout the 1000-second test interval. This demonstrates that the implementation of TDMA-based protection for mission-critical traffic maintains compatibility with existing robotic data exchanges without degrading their operational performance.

The Fig. \ref{fig_tdma_sequence_multi}(c) shows the delivery status of mission-critical control packets. The definition of delivery status is the same as in Section \ref{sec-II-B}. With the hybrid TDMA/CSMA protocol, the missed deadline error rate is reduced by 93\% compared to the conventional CSMA baseline. In a 1000-second experimental evaluation conducted under identical network conditions as the CSMA baseline, only 0.51\% of packets experienced packet loss error or missed deadline error. Detailed error analysis shows that packet loss errors account for 98\% of total failures, while missed deadline errors constitute only a negligible fraction of the total  errors. Notably, the transmission errors in mission-critical control traffic in hybrid protocol are sparse and temporally isolated. This is different from the case in CSMA where missed deadline errors tend to occur in bursts due to contention and unpredictable queuing delays.

To evaluate the practical impact of the proposed hybrid protocol, we ran a robot path-tracking simulation in ROS environment  \cite{macenski2022robot} using the PurePursuit trajectory following algorithm \cite{rokonuzzaman2021review}. We design a simulation framework to compare the CSMA communication baseline with the proposed hybrid protocol in robot coordination scenarios.  

The simulation framework is implemented as a modular ROS architecture consisting of three components: (i) a trajectory generation module, (ii) a PurePursuit control module, and (iii) a communication emulation layer. The trajectory module uses cubic-spline interpolation of sinusoidal waypoints to generate smooth reference paths and emits continuous position and orientation references at 1-meter intervals. The PurePursuit control module implements the geometric path-following algorithm with a velocity-adaptive lookahead to compute steering angles from the geometric relationship between the current pose and the target lookahead point on the reference trajectory. Control commands are generated at a nominal frequency of 10 Hz for responsive trajectory tracking. The communication emulation layer injects the packet transmission sequences from our real-time SDR testbed as timing constraints into the ROS control loop. When communication failures occur, the system maintains the previous steering command while logging the missed deadline.

\begin{figure}[htbp]
    \centering
    \begin{subfigure}[t]{0.15\textwidth}
        \centering
        \includegraphics[width=\textwidth]{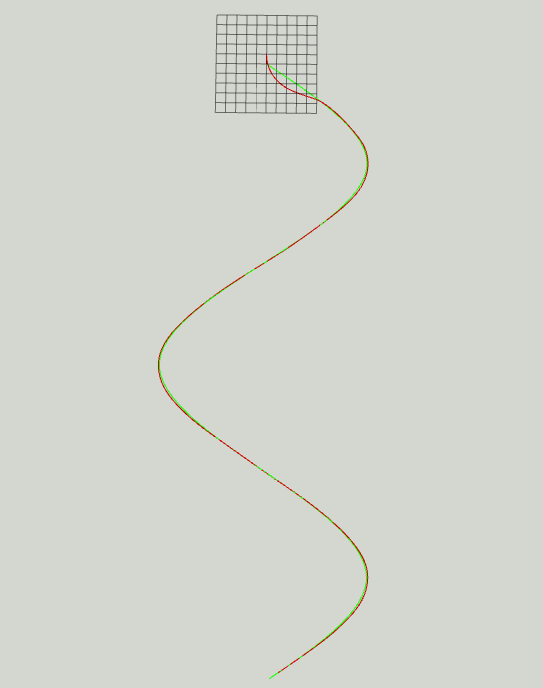}
        \caption{CSMA, v=3 m/s}
        \captionsetup{font={small}}
        \label{fig_ros_rviz_demo_csma_v3}
    \end{subfigure}
    \hfill
    \begin{subfigure}[t]{0.15\textwidth}
        \centering
        \includegraphics[width=\textwidth]{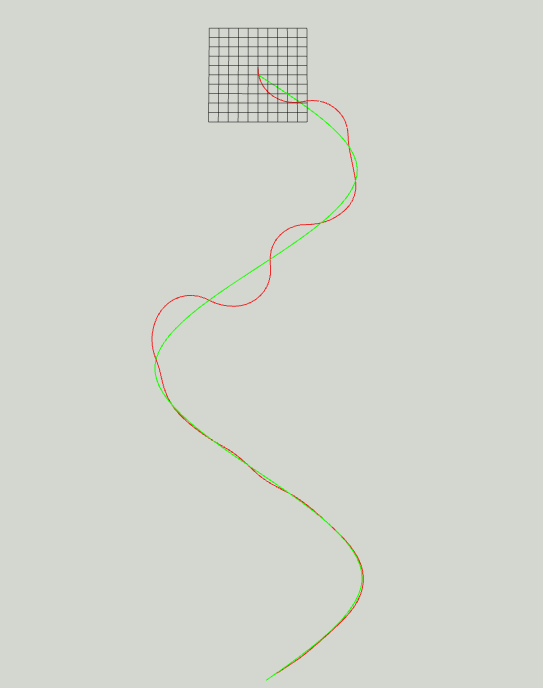}
        \caption{CSMA, v=6 m/s}
        \captionsetup{font={small}}
        \label{fig_ros_rviz_demo_csma_v6}
    \end{subfigure}
    \hfill
    \begin{subfigure}[t]{0.15\textwidth}
        \centering
        \includegraphics[width=\textwidth]{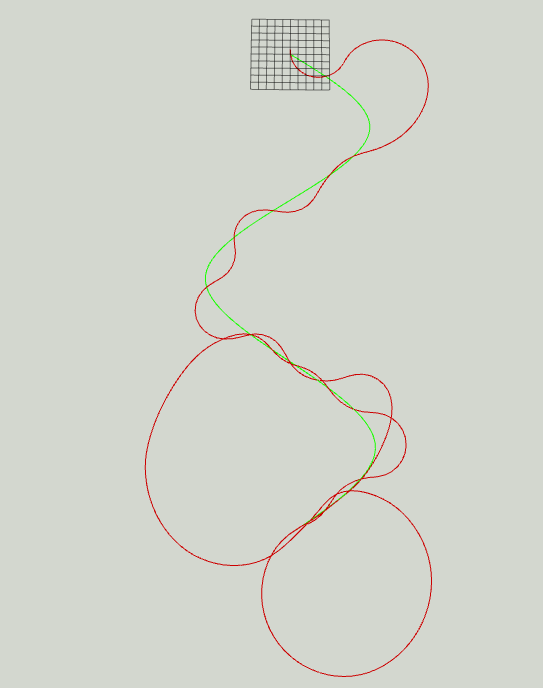}
        \caption{CSMA, v=9 m/s}
        \captionsetup{font={small}}
        \label{fig_ros_rviz_demo_csma_v9}
    \end{subfigure}
    
    \vspace{0.5em} 

    \begin{subfigure}[t]{0.15\textwidth}
        \centering
        \includegraphics[width=\textwidth]{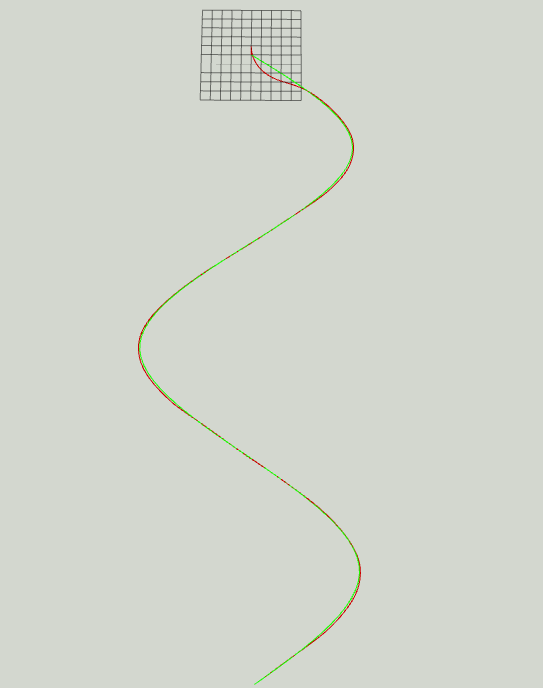}
        \caption{Hybrid, v=3 m/s}
        \captionsetup{font={small}}
        \label{fig_ros_rviz_demo_hybrid_v3}
    \end{subfigure}
    \hfill
    \begin{subfigure}[t]{0.15\textwidth}
        \centering
        \includegraphics[width=\textwidth]{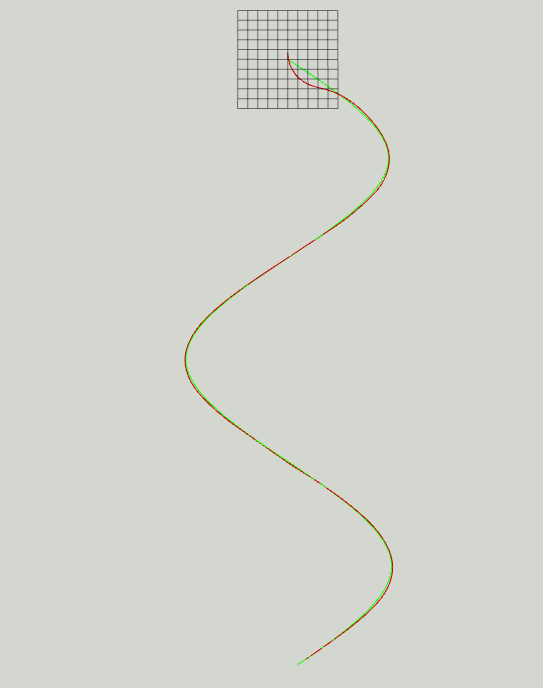}
        \caption{Hybrid, v=6 m/s}
        \captionsetup{font={small}}
        \label{fig_ros_rviz_demo_hybrid_v6}
    \end{subfigure}
    \hfill
    \begin{subfigure}[t]{0.15\textwidth}
        \centering
        \includegraphics[width=\textwidth]{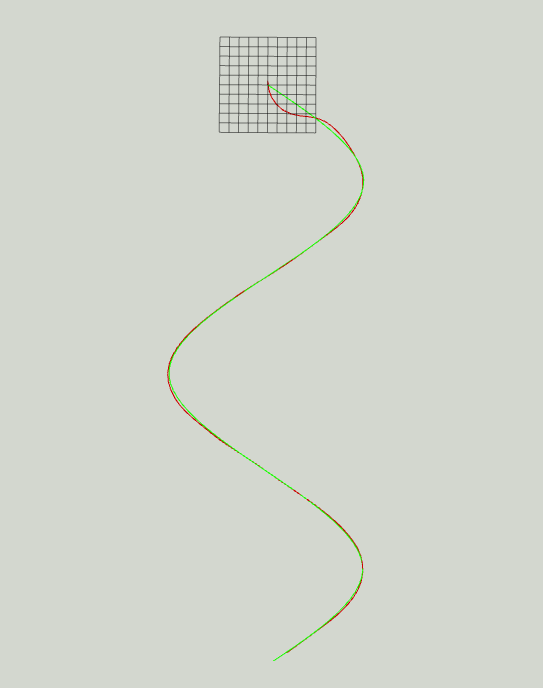}
        \caption{Hybrid, v=9 m/s}
        \captionsetup{font={small}}
        \label{fig_ros_rviz_demo_hybrid_v9}
    \end{subfigure}
    \captionsetup{font={small}}
    \caption{Robot trajectories (red) versus reference path (green) at 3 m/s, 6 m/s, and 9 m/s under CSMA, and the proposed hybrid TDMA/CSMA protocol (Hybrid)}
    \label{fig_ros_rviz_demo}
\end{figure}

We ran the robot simulation in the Gazebo physics simulator \cite{koenig2004design} using a differential drive robot model with odometry sensors and kinematic constraints. The simulation environment includes realistic friction coefficients and inertial properties to match real robotic platforms. The robot's initial position is set at the trajectory starting point. The simulator continuously logged performance metrics, including lateral tracking error, longitudinal position error, heading deviation, and control command execution statistics. We used RViz’s real-time visualization to present an intuitive trajectory comparison by overlaying the tracked and reference paths as shown in Fig. \ref{fig_ros_rviz_demo}.

Fig. \ref{fig_ros_rviz_demo} presents the predefined reference trajectory together with the robot’s tracking path under both the CSMA baseline and the proposed hybrid TDMA/CSMA framework. In the CSMA baseline, the actual tracking path shows notable deviations from the reference, with the largest errors occurring in curved sections that demand rapid and precise steering command updates. Contention‑based CSMA causes mission‑critical packets to miss their deadlines and leads to bursty errors that delay multiple consecutive commands. This effect increases in turning segments, where steering commands require more frequent and time‑sensitive updates. As a result, the robot continues to execute outdated steering angles for extended periods, causing position errors to accumulate. In severe cases as shown in Fig. \ref{fig_ros_rviz_demo_csma_v6} and Fig. \ref{fig_ros_rviz_demo_csma_v9}, the tracking trajectory forms large arcs or unintended loops before updated commands are received, resulting in severe tracking errors.

In contrast, slot-based TDMA greatly reduces queuing delays from competing traffic and guarantees that mission-critical traffic arrives within their deadlines. Moreover, the TDMA slot allocation of the proposed framework can be dynamically adjusted according to the real‑time requirements of the robot’s application to provide adequate channel resources. As a result, the tracking path in the hybrid case remains closely aligned with the reference across both straight and curved sections, with small lateral error and smooth transitions through turns.

To quantify the path‑tracking performance of the robots based on the deviation between the predefined reference trajectory and the robot’s actual tracking path, we first define the instantaneous lateral tracking error. Let the reference trajectory be a sequence of sampled points

\begin{equation}
\mathbf{r}_k = [x_r(k), y_r(k)]^{\top}, \quad k = 1,2,\dots,N , 
\end{equation}

where $N$ is the total number of reference samples and ${{x}_{r}}(k),{{y}_{r}}(k)$ are the Cartesian coordinates of the $k$-th reference point in the global (world) coordinate frame.  
The actual robot trajectory obtained from ROS is given by discrete position samples

\begin{equation}
\mathbf{p}_i = [x(i), y(i)]^{\top}, \quad i = 1,2,\dots,M ,   
\end{equation}

where $M$ is the total number of recorded robot positions and $x(i),y(i)$ are the Cartesian coordinates of the robot position at time $i$ in the same global coordinate frame.
For each robot position $\mathbf{p}_i$, its nearest point $\mathbf{r}_k$ on the reference trajectory is defined by

\begin{equation}
\mathbf{r}_k = \arg\min_{\mathbf{r} \in \{\mathbf{r}_1, \mathbf{r}_2, \dots, \mathbf{r}_N\}} \| \mathbf{p}_i - \mathbf{r} \|.  
\end{equation}

The instantaneous lateral error ${{e}_{d}}(i)$ is then defined as:

\begin{equation}
{{e}_{d}}(i)=\|{{\mathbf{p}}_{i}}-{{\mathbf{r}}_{k}}\|\sin {{\theta }_{i}}.
\end{equation}

where $\theta_i$ is the angle between the deviation vector $(\mathbf{p}_i - \mathbf{r}_k)$ and the tangent vector $\mathbf{t}_k$ at $\mathbf{r}_k$. 

\begin{figure}[htbp]
\centerline{\includegraphics[width=\linewidth]{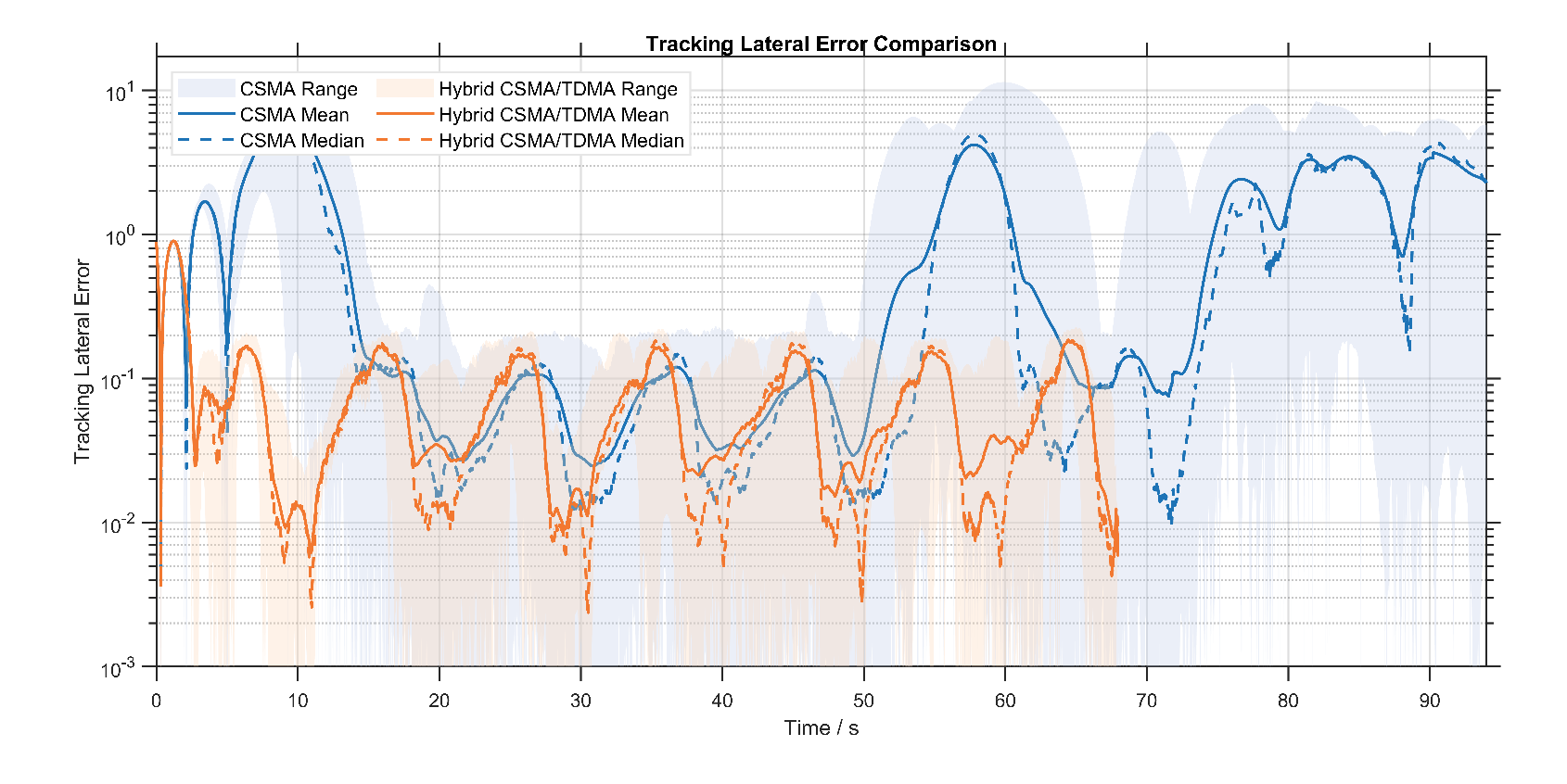}}
\centering
\captionsetup{font={small}}
\caption{Comparison of Lateral Errors for CSMA and Hybrid TDMA/CSMA at 3 m/s}
\label{fig_lateral_error_comparison_v3}
\end{figure}

\begin{figure}[htbp]
\centerline{\includegraphics[width=\linewidth]{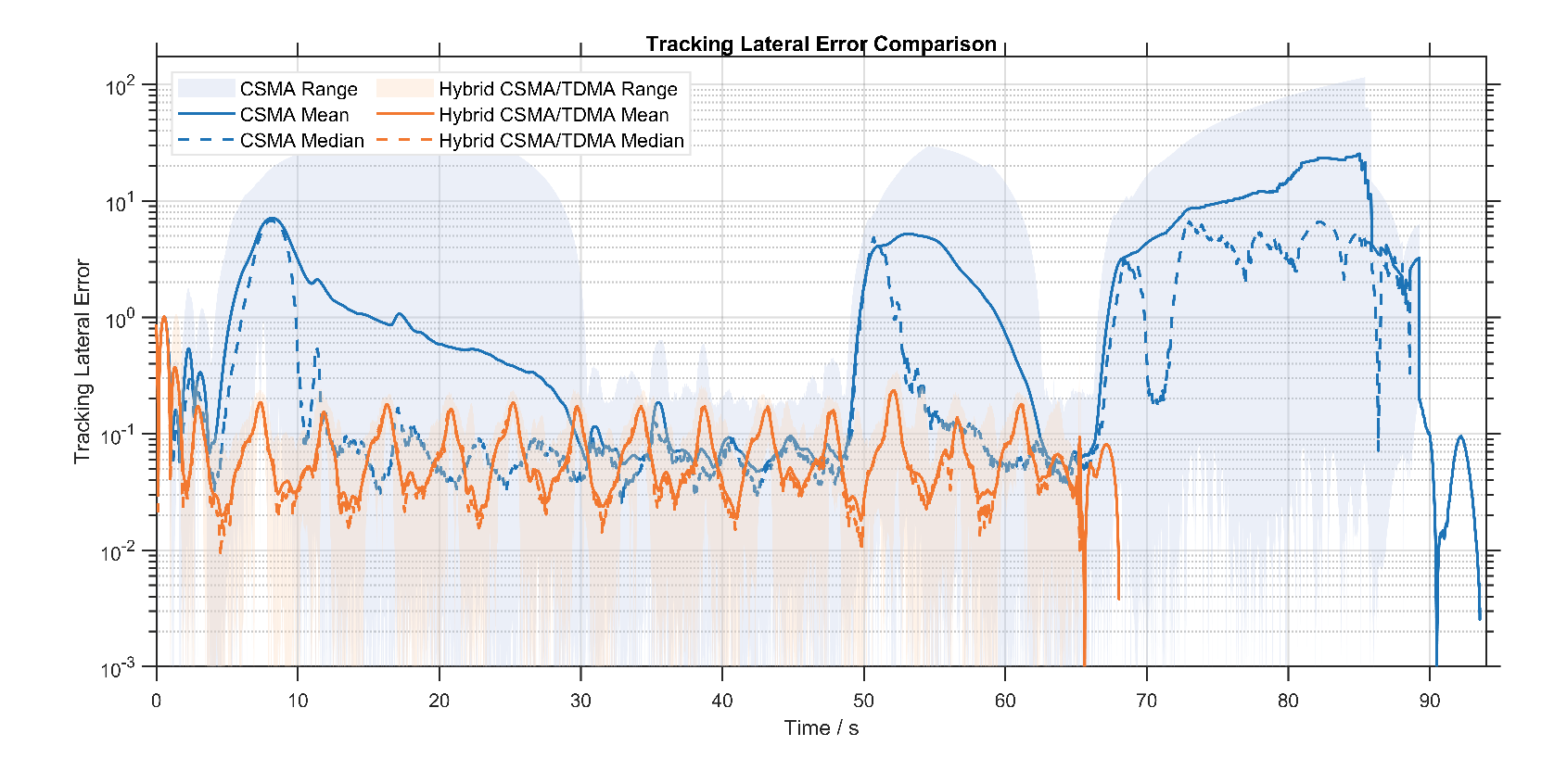}}
\centering
\captionsetup{font={small}}
\caption{Comparison of Lateral Errors for CSMA and Hybrid TDMA/CSMA at 6 m/s}
\label{fig_lateral_error_comparison_v6}
\end{figure}

\begin{figure}[htbp]
\centerline{\includegraphics[width=\linewidth]{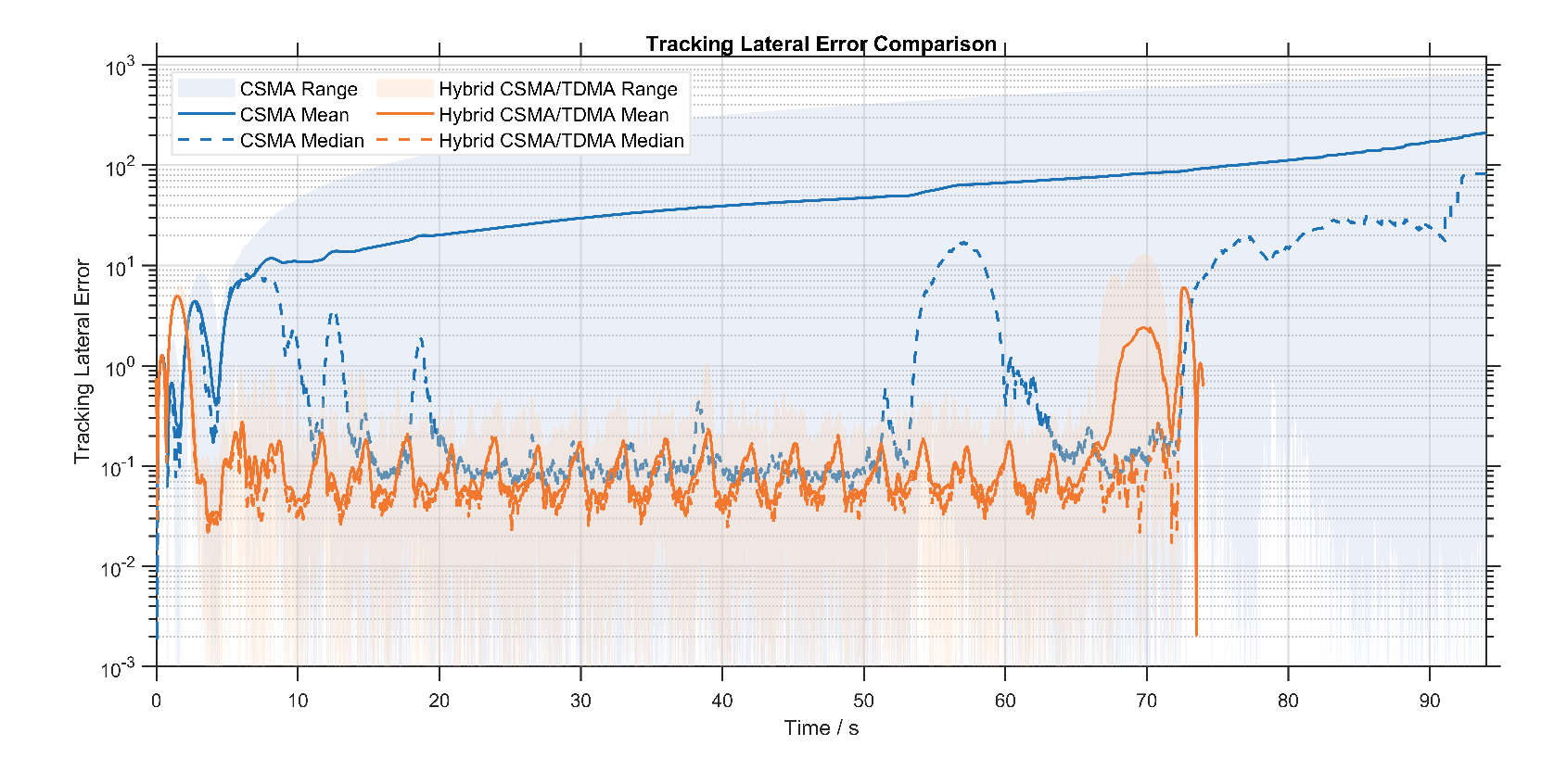}}
\centering
\captionsetup{font={small}}
\caption{Comparison of Lateral Errors for CSMA and Hybrid TDMA/CSMA at 9 m/s}
\label{fig_lateral_error_comparison_v9}
\end{figure}

To evaluate the performance of the proposed hybrid TDMA/CSMA framework, we perform the path‑tracking simulation in a ROS environment that emulate real‑world navigation scenarios under varying speeds and network conditions. Fig. \ref{fig_lateral_error_comparison_v3}, \ref{fig_lateral_error_comparison_v6}, \ref{fig_lateral_error_comparison_v9} show the absolute lateral tracking errors at different speeds for the CSMA baseline and the proposed hybrid TDMA/CSMA  framework. Each scenario was repeated 100 times to observe the distribution of lateral errors with variations from the ROS Gazebo physics engine's stochastic effects.

At lower speed (Fig. \ref{fig_lateral_error_comparison_v3}), CSMA baseline shows occasional lateral error bursts exceeding 1 m, mainly in curved sections where deadline misses occur in clusters. The mean lateral error remains below 0.3 m for most of the run but spikes to the 10 m during miss-deadline command loss. The hybrid protocol avoids these bursts and keeps both mean and median errors within 0.1 m across the whole trajectory. Initial deviations result from eliminating ROS starting point offsets. At moderate (Fig. \ref{fig_lateral_error_comparison_v6}) and high speed (Fig. \ref{fig_lateral_error_comparison_v9}), CSMA baseline shows worsening burst peaks lateral error, exceeding 10 m  at v=6 m/s and surpassing 20 m toward the end at v = 9 m/s. The rising median errors are caused by more frequent deadline misses and the prolonged application of outdated commands. In contrast, hybrid TDMA/CSMA protocol maintains stable performance across both cases, keeping mean errors below 0.2 m, median errors near 0.1 m, and rapidly correcting transient deviations to prevent drift.

In addition to lateral deviations, we also consider the root‑mean‑square (RMS) trajectory error, which characterizes the overall deviation between the actual and reference trajectories. Since $M$  and $N$  may differ due to different sampling rates, the reference trajectory is linearly interpolated to the timestamps of the actual trajectory. The RMS trajectory error ${{e}_{\text{traj,RMS}}}$ is then defined as

\begin{equation}
{{e}_{\text{traj,RMS}}}=\sqrt{\frac{1}{M}\sum\limits_{i=1}^{M}{\|{{\mathbf{p}}_{i}}-{{\widetilde{\mathbf{r}}}_{i}}{{\|}^{2}}}}.    
\end{equation}

where $\tilde{\mathbf{r}}_i$ is the interpolated reference position corresponding to $\mathbf{p}_i$.

\begin{figure}[htbp]
\centerline{\includegraphics[width=\linewidth]{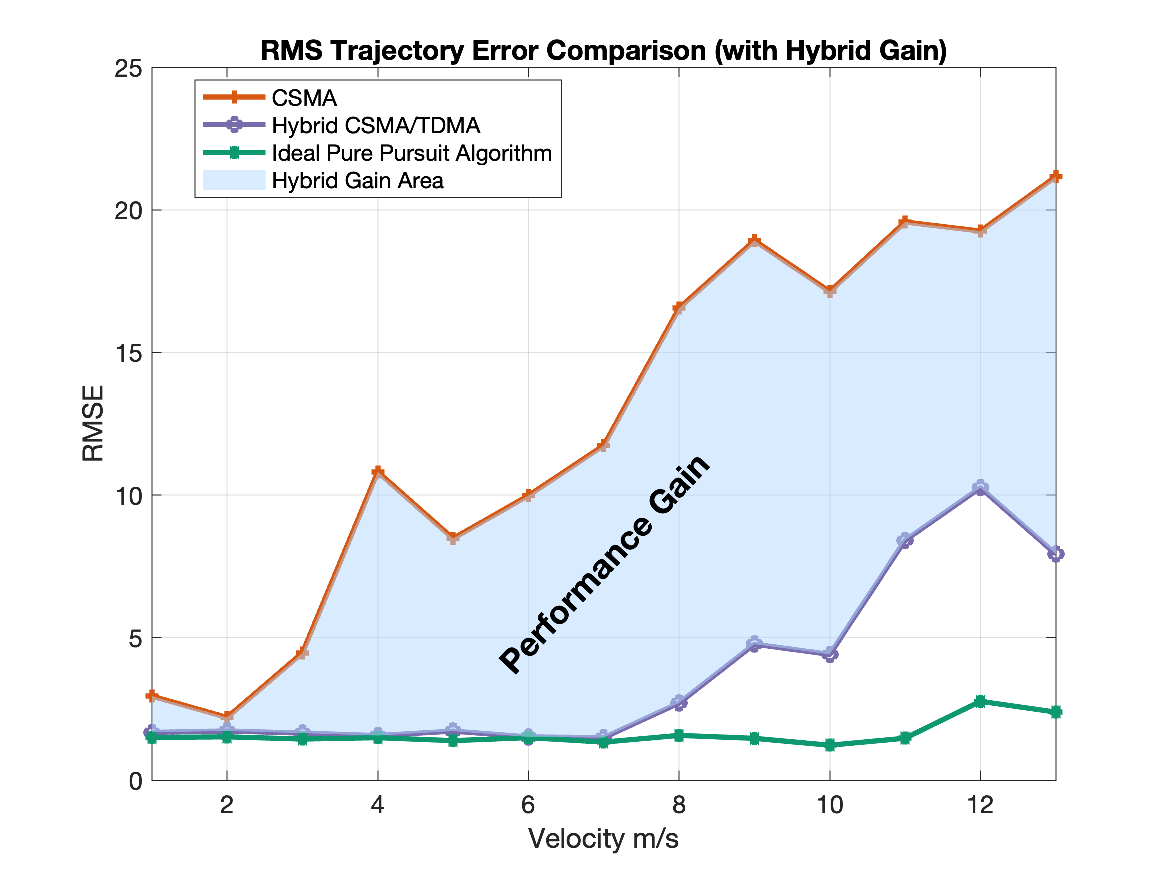}}
\centering
\captionsetup{font={small}}
\caption{RMS trajectory error comparison under different velocity}
\label{fig_rmse_demo}
\end{figure}


Fig. \ref{fig_rmse_demo} shows the RMS trajectory error of the robotic path-tracking experiment at different velocities. The results compare the proposed hybrid TDMA/CSMA framework, the CSMA baseline, and an ideal PurePursuit algorithm baseline. The PurePursuit baseline represents an ideal path‑tracking controller without any communication delay and error. In this case, the system maintains minimal RMS error and closely follows the reference trajectory at all tested velocities. When command losses or missed deadlines are introduced by the communication emulation layer, the CSMA baseline shows a rapid RMS error increase as velocity rises, reaching about 25 m at 12 m/s. The hybrid TDMA/CSMA framework maintains low RMS error, close to the ideal baseline, because it avoids burst sequences of consecutive missed deadlines and only incurs occasional errors caused by poor channel conditions. Fig. \ref{fig_rmse_demo} also shows that the performance gain of the hybrid protocol is greater at higher velocities. In these cases, control commands require tighter timing to avoid cumulative trajectory deviations because a single error can trigger unrecoverable path‑tracking failures.

The path‑tracking evaluation provides a representative case for comparing the CSMA baseline and the gybrid TDMA/CSMA framework, as similar requirements exist in many robotic applications. Systems such as surgical robots demand precise and timely control signals to execute fine movements, while drone swarms rely on synchronized command delivery to maintain coordinated flight. In both cases, delays or bursts of packet loss can cause severe and irreversible errors. By prioritizing time‑sensitive traffic and limiting burst losses, the hybrid TDMA/CSMA framework enhances control accuracy and operational stability, making it broadly applicable to demanding real‑time robotic scenarios.

\section{Conclusion and Future Work}\label{sec-V}
In this work, we first analyze robotic communication patterns, focusing on the deadline‑constrained communication in mission‑critical control and the dynamic robotic traffic. Real-time SDR emulation experiments show that conventional CSMA cannot meet stringent and dynamic timing requirements when heterogeneous robot traffic shares the same medium, leading to bursty and consecutive missed deadlines error even without actual packet loss.

Motivated by these observations, we present a hybrid TDMA/CSMA protocol to address command‑traffic congestion in time‑sensitive robotic applications. By combining TDMA’s deterministic slot scheduling with CSMA’s flexible access, the protocol guarantees collision‑free, low‑latency delivery of mission‑critical commands while supporting heterogeneous robotic traffic. The proposed protocol has three key features: (i) high‑precision PTP synchronization for sub‑microsecond slot alignment, (ii) a three‑section superframe with dynamic TDMA slot allocation based on real‑time network conditions and application demands, and (iii) beacon‑based NAV protection to reserve the medium for TDMA and control sections against external interference. The proposed protocol also supports seamless integration with IEEE 802.11‑based wireless infrastructures for cost‑effective deployment and rapid scaling in heterogeneous robotic networks. 

The proposed protocol implemented on a real-time Zynq‑7000 SoC‑based SDR platform and the experimental data collected from a real wireless environment were tested in ROS using the PurePursuit tracking algorithm. In high‑speed path tracking, it reduced RMS trajectory error by up to 90\% and kept packet loss for latency‑sensitive control messages near zero compared to conventional CSMA baseline. It is also applicable to general real‑time robot control and various cooperative robotic systems that require precise and time‑critical command delivery.

Future work will focus on cross‑layer design to enhance the protocol and on comprehensive real‑world robotic validation. First, we aim to achieve integration with ROS network architectures by designing unified communication interfaces and optimizing message‑passing mechanisms. Second, we aim to implement the proposed framework on physical robotic platforms like mobile robots and robotic arms to evaluate the protocol’s performance under real‑world conditions. Third, we plan to develop deadline-aware scheduling algorithms on various communication layer to dynamically adjust TDMA slot allocations based on real-time traffic demands. Collectively, these efforts are expected to enhance the framework’s robustness, scalability, and applicability in AI‑driven, real‑time robotic control and support the development of safer and more efficient autonomous robot system.

\bibliographystyle{IEEEtran}
\bibliography{./Reference/Ref_IEEE}

\end{document}